\documentclass[runningheads]{llncs}

\usepackage{eccv}

\usepackage{eccvabbrv}
\usepackage{mdframed}
\usepackage{amsmath}

\usepackage{graphicx}
\usepackage{booktabs}
\definecolor{Gray}{gray}{0.9}
\usepackage[accsupp]{axessibility}  

\usepackage{amsfonts}
\usepackage{amssymb}
\usepackage{bbding}
\usepackage{multirow}
\usepackage{rotating}
\usepackage{algorithm} 
\usepackage{adjustbox}
\usepackage{color}
\usepackage{epsfig}
\usepackage{graphicx}
\usepackage{amsmath}
\usepackage{booktabs}
\usepackage{colortbl}
\usepackage{algorithm}
\usepackage{algorithmic}
\usepackage{newfloat}
\usepackage{listings}
\usepackage{soul}
\usepackage{wrapfig}

\usepackage{orcidlink}
\usepackage{bbding}

\begin{document}


\title{Stripe Observation Guided Inference Cost-free Attention Mechanism}

\titlerunning{Stripe Observation Guided
Attention Mechanism}

\author{Zhongzhan Huang $^*$\inst{1}\orcidlink{0000-0002-2135-1104} \and
Shanshan Zhong \thanks{Equal contributions} \inst{1}\orcidlink{0000-0003-3082-7351} \and
Wushao Wen\inst{1}\orcidlink{0000-0003-4819-4679}
\and \\
Jinghui Qin\inst{2}\orcidlink{0000-0003-0663-199X}
\and
Liang Lin\inst{1,3}\orcidlink{0000-0003-2248-3755} 
}

\authorrunning{Zhongzhan Huang et al.}

\institute{ Sun Yat-sen University \and Guangdong University of Technology \and
 Peng Cheng Laboratory\\
\email{ \{huangzhzh23,zhongshsh5\}@mail2.sysu.edu.cn}\\
Correspondence: \email{linliang@ieee.org}\\
}

\maketitle

\begin{abstract}

Structural re-parameterization (SRP) is a novel technique series that boosts neural networks without introducing any computational costs in inference stage.
 The existing SRP methods have successfully considered many architectures, such as normalizations, convolutions, etc. However, the widely used  but computationally expensive attention modules cannot be directly implemented by SRP due to the inherent multiplicative manner and the modules' output is input-dependent during inference.  In this paper, we statistically discover a counter-intuitive phenomenon \textit{Stripe Observation} in various settings, which reveals that channel attention values consistently approach some constant vectors during training. It inspires us to propose a novel attention-alike SRP, called ASR, that allows us to achieve SRP for a given network while enjoying the effectiveness of the attention mechanism. Extensive experiments conducted on several standard benchmarks show the effectiveness of ASR in generally improving the performance of various scenarios without any elaborated model crafting. 
 We also provide experimental evidence for how the proposed ASR can enhance model performance. \url{https://github.com/zhongshsh/ASR}.
\end{abstract}

\section{Introduction}
\label{sec:intro}

\begin{table}[htbp]
  \centering
  \caption{The significant decrease in inference speed (Frames Per Second) by using attention modules. "\#P." denotes the number of parameters.}
  \resizebox{0.99\hsize}{!}{
    \begin{tabular}{lcccc|r|lcccc}
    \toprule
          & \multicolumn{2}{c}{CIFAR100} & \multicolumn{2}{c|}{STL10} &       &       & \multicolumn{2}{c}{CIFAR100} & \multicolumn{2}{c}{STL10} \\
\cmidrule{2-5}\cmidrule{8-11}    Model & \multicolumn{1}{c}{\#P. (M)} & \multicolumn{1}{c}{Speed} & \multicolumn{1}{c}{\#P. (M)} & \multicolumn{1}{c|}{Speed} &       & Model & \multicolumn{1}{c}{\#P. (M)} & \multicolumn{1}{c}{Speed} & \multicolumn{1}{c}{\#P. (M)} & \multicolumn{1}{c}{Speed} \\
    \midrule
    \rowcolor{Gray} ResNet164 & 1.73 &   1944    &  1.70     &  255     &       & ResNet164 & 1.73 &   1944    &  1.70     &  255  \\
    +IE~\cite{wang2021recurrent,liang2020instance}  & 1.74  & 1505 ({\color{teal}{$\downarrow$} 22.56\%})  & 1.72  & 170 ({\color{teal}{$\downarrow$} 33.26\%})    &    & +SRM~\cite{lee2019srm} & 1.76  & 1387 ({\color{teal}{$\downarrow$} 28.64\%})  & 1.74  & 162 ({\color{teal}{$\downarrow$} 36.40\%})  \\
    +CBAM~\cite{woo2018cbam} & 1.93  & \ \ 793 ({\color{teal}{$\downarrow$} 59.21\%})   & 1.90  & 127 ({\color{teal}{$\downarrow$} 50.23\%})       &       & +DIA~\cite{huang2020dianet} & 1.95  & 1092 ({\color{teal}{$\downarrow$} 43.82\%})  & 1.92  & 154 ({\color{teal}{$\downarrow$} 39.61\%})  \\
    +SE~\cite{hu2018squeeze} & 1.93  & 1469 ({\color{teal}{$\downarrow$} 24.42\%})  & 1.91  & 173 ({\color{teal}{$\downarrow$} 32.08\%})      &       & +SPA~\cite{guo2020spanet} & 3.86  & 1080 ({\color{teal}{$\downarrow$} 44.45\%})  & 3.83  & 180 ({\color{teal}{$\downarrow$} 29.36\%}) \\
    \rowcolor{Gray} +ASR (SE) & 1.73  & 1942 ({\color{red}{$\sim$} 0.00\%})  & 1.70  & 255 ({\color{red}{$\sim$} 0.00\%})       &       & +ASR (SPA) & 1.73  & 1946 ({\color{red}{$\sim$} 0.00\%})  & 1.70  & 253 ({\color{red}{$\sim$} 0.00\%})  \\
    \bottomrule
    \end{tabular}%
    }
  \label{tab:fps}%
\end{table}%

\begin{figure*}[t]
  \centering
 \includegraphics[width=\linewidth]{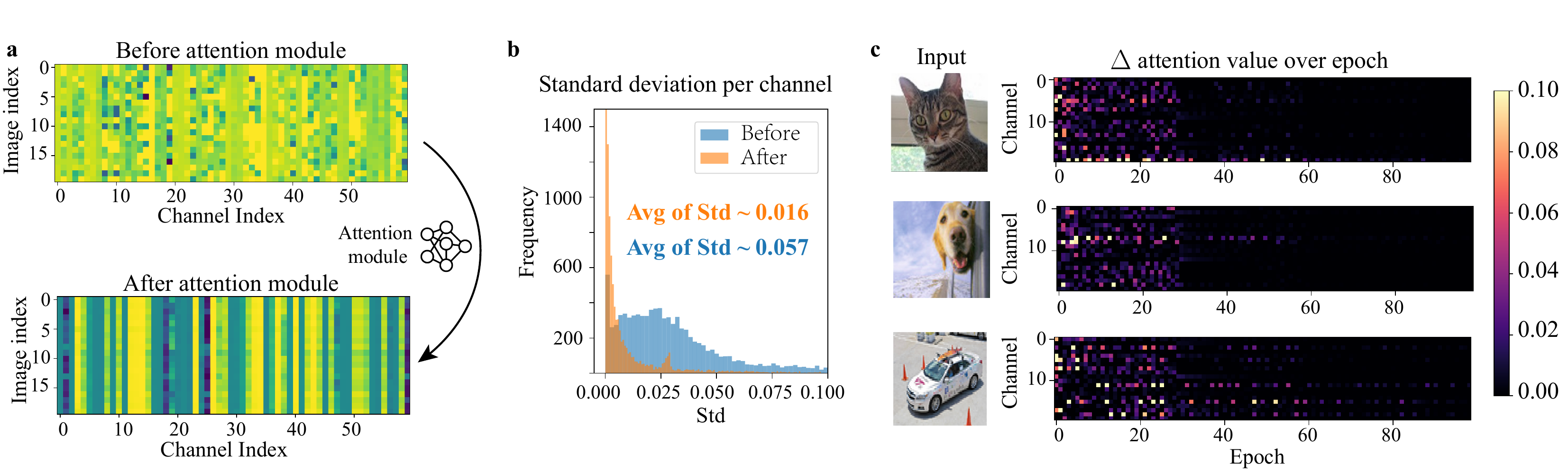}
  \caption{The visualization of the \textit{Stripe Observation} on ImageNet classification. In Section \ref{sec:settings}, we shows this observation also holds for segmentation and detection tasks. This phenomenon can be verified on a large number of third-party models with different settings, ensuring its \ul{reproducibility and generality}. (\textbf{see \ul{Appendix} for comprehensive examples})  \textbf{a}, after passing through the attention model, the attention values of different images tend to approach a certain value within the same channel, resulting in a "stripe structure". \textbf{b}, \textbf{Statistical analysis}. The standard deviation of attention values for each channel is almost distributed around zero. \textbf{c}, the visualization of the first-order difference (absolute value) for attention value over epoch. Most of the values approach some constants rapidly and consistently. }
\label{fig:main}
\end{figure*}
\begin{figure*}[t]

   \centering
\includegraphics[width=0.8\linewidth]{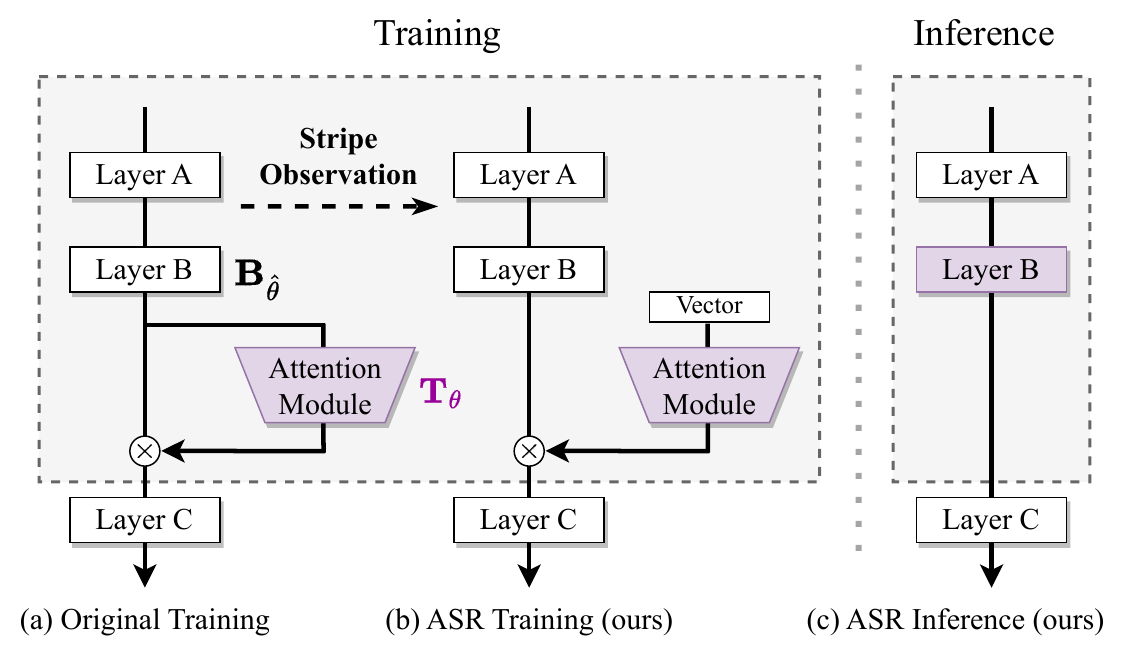}
  \caption{The sketch of ASR. Inspired by \textit{Stripe Observation}, we utilize a learnable vector as input to the attention module for training. In Inference phase, the attention module can be merged into the parameters of the backbone. }
\label{fig:struc}
\end{figure*}

The structural re-parameterization (SRP) technique \cite{Guo20,Cao20,ding2021repvgg,hu2022online,wang2022repsr} is a type of efficient neural network technology that decouples training and inference, greatly facilitating the deployment of Deep Neural Networks (DNNs) in practical applications and possessing excellent potential for industrial implementation. It increases the model's representation power for a given backbone network by adding multiple branches or specific layers with various neural network components to the backbone during the training phase, while during the inference phase, the added branches or layers can be merged into the parameters of the backbone network through some equivalent transformations, enabling performance improvement without any additional parameters or computational costs. 
Specifically,
for an input $x$, a branch layer $\mathbf{T}_\theta$ with learnable parameters $\theta$, and a layer on the backbone $\mathbf{B}_{\hat{\theta}}$ (usually a convolutional filter with a sufficiently large kernel size \cite{ding2021diverse}), if there are transformations $h$ and $g$ such that

\begin{equation}
\underbrace{h[\mathbf{T}_{\theta},\mathbf{B}_{\hat{\theta}}](x) }_{\text{Training}} = \underbrace{ \mathbf{B}_{g[\hat{\theta},\theta]} (x) }_{\text{Inference}},
    \label{eq:ccs}
\end{equation}

this situation can be decoupled in training and inference processing using SRP. For example, during training, $\mathbf{Conv}_{\theta}(x) + \mathbf{Conv}_{\hat{\theta}}(x)$ can be equivalently transformed into $\mathbf{Conv}_{\theta+\hat{\theta}}(x)$ during inference, enabling the model performance of two convolutional filters to be achieved with only one convolutional filter. Another example about batch normalization (BN) is $\mathbf{BN}_{\theta}(\mathbf{Conv}_{\hat{\theta}}(x))$ can be equivalently transformed into $\mathbf{Conv}_{\hat{\theta},a,b}(x)$ during inference, where $a$ and $b$ are constants related to $\theta$. Based on the idea of Eq.~(\ref{eq:ccs}), although previous SRP methods have successfully integrated various neural network components, including normalization methods, multi-branch convolution, global pooling, etc.,
existing methods are currently unable to be applied to attention modules, which are widely used in deep learning applications.
This is due to the fact that the attention module \textcolor[RGB]{169,38,217}{$\mathbf{T}_{\theta}$ }, as illustrated in Fig.~\ref{fig:struc} (a), acts on the backbone network with an inherent multiplicative manner~\cite{fu2019dual,hou2021coordinate,gao2019global,huangattns,zhong2023lsas,huang2023scalelong,cao2019gcnet,liang2022balancing,zhang2020resnest,qin2021fcanet} (usually is the element-wise multiplication $\odot$), and the module's output is also input-dependent during inference. Therefore, usually, for any given transformation $g$, we have
\begin{equation}
\mathbf{B}_{\hat{\theta}} \odot \textcolor[RGB]{169,38,217}{\mathbf{T}_{\theta} } \neq \mathbf{B}_{g[\hat{\theta},\textcolor[RGB]{169,38,217}{\theta}]}. 
    \label{eq:attno}
\end{equation}
Eq.~(\ref{eq:attno}) reveals that the attention module is not directly compatible with the existing SRP's paradigm, which means that the widely used attention networks may limit the scenarios and scope of SRP. However, attention modules have an unignorable impact on inference. Despite their few parameters, as shown in Table~\ref{tab:fps}, attention modules can drastically slow inference speed due to complex computations. 

Therefore, there is a question: \textit{\textbf{Can we incorporate the effectiveness of attention mechanisms into SRP in an indirect way? }}To answer this question, we first find and verify  a counter-intuitive  phenomenon that occurs in different datasets, network structures, and attention modules, that is

\begin{mdframed}[backgroundcolor=gray!8]
\begin{minipage}{\linewidth}
\noindent \textbf{(\textit{Stripe Observation})} As shown in Fig.~\ref{fig:main}, the attention value $\mathbf{v}_i\in \mathbb{R}^c$ obtained by $i$-th channel attention module will consistently approach to a value $\mathbf{\bar{v}}_i\in \mathbb{R}^c$, which $\mathbf{\bar{v}}_i$ may follow a distribution with mean vector $\mathbf{\mu}_i$ and covariance matrix $\mathbf{diag}(\sigma_{i1}^2,...,\sigma_{ic}^2)$,  $\sigma_{ij}, j = 1,2,...,c$ closes to 0. 

\end{minipage}
\end{mdframed}

Empirically, as shown in Fig.~\ref{fig:main}, we visualize the attention values of the well-known attention module SE~\cite{hu2018squeeze} on the ResNet50~\cite{he2016deep} architecture trained on the ImageNet~\cite{russakovsky2015imagenet} dataset to observe the so-called \textit{Stripe Observation}. For Fig.~\ref{fig:main} \textbf{(a)}, we consider the images from the ImageNet dataset as inputs to the model and visualize the values of feature maps before and after passing through the attention module. We find that the values of the input feature maps of the attention module varied greatly among different channels and images, while the output values of different images generated from the attention module tended to approach a certain value within the same channel, resulting in a "stripe structure". Therefore, we call this phenomenon \textit{Stripe Observation}. 

Further, we can quantify this observation from a statistical perspective. Specifically, we visualize the standard deviation of values for each channel of the feature map before and after passing through the attention modules, as shown in Fig.~\ref{fig:main} \textbf{(b)}. We observe that the standard deviation of values for each channel is almost distributed around zero after passing through the attention module, regardless of the images, indicating that the attention values at any module do indeed approach a corresponding constant vector. Let $\mathbf{v}_i^t$ be the attention value obtained by the $i$-th module during the $t$-th epoch, We can further explore the process of forming these constant vectors. Fig.~\ref{fig:main} \textbf{(c)} visualizes the first-order difference (absolute value) of three input images over epoch for randomly selected 20 channels, i.e., $\Delta_i = \textbf{abs}[\mathbf{v}_i^{t+1} - \mathbf{v}_i^t]$. Typically, a commonly used schedule learning rate is applied for training ResNet50 on the ImageNet dataset, where the learning rate is reduced by 90\% every 30 epochs. As shown in Fig.~\ref{fig:main} \textbf{(c)}, most of the values almost converge at the first learning rate decay (30 epochs), i.e., the first-order difference is zero (black color). More empirical visualizations about \textit{Stripe Observation} can be found in the \ul{Appendix}.

Inspired by the \textit{Stripe Observation} that the channel attention values of different inputs in a dataset tend to approach a constant vector, we propose an Attention-alike Structural Re-parameterization (ASR) as shown in Fig.~\ref{fig:struc} (b), where we directly consider a learnable vector as the input to the attention modules. This makes the attention modules relatively independent of the model input so that the "attention values" become a constant vector after training directly. Therefore, the dilemma mentioned in Eq.~(\ref{eq:attno}) is solved, and we can achieve the SRP, like Fig.~\ref{fig:struc} (c), of a given model while enjoying the effectiveness of the attention mechanism. The details of ASR can be found in Section \ref{sec:method}. We also demonstrate the effectiveness of ASR and its compatibility with existing neural network training methods using multiple benchmarks in Section \ref{sec:experiment}. Furthermore, we provide the empirical evidence for the strong robustness of ASR in Section \ref{sec:analysis}. We summarize contributions as follows:
\begin{itemize}
     \item We revisit the channel attention mechanism from a statistical perspective. With comprehensive experiments, we find the \textit{Stripe Observation}, which reveals that the attention value approach constant vectors during training.
     \item We coin a novel structural re-parameterization method ASR, tailored for \textit{Stripe Observation}, which can leverage various attention modules to improve the model without any extra inference cost.
     \item We provide experimental evidence for the proposed ASR's compatibility, effectiveness, and robustness. 
\end{itemize}

\section{Related works}

\noindent{\textbf{Attention mechanism}} selectively focuses on the most informative components of a network via self-information processing and has gained a promising performance on vision tasks \cite{liang2020instance}. 
For example, SENet \cite{hu2018squeeze} proposes the channel attention mechanism, which adjusts the feature map with channel view, and CBAM \cite{woo2018cbam} considers both channel and spatial attention for adaptive feature refinement. 
Recently, more works \cite{fu2019dual,hou2021coordinate,zhu2019empirical,gao2019global,cao2019gcnet,liang2022balancing,zhang2020resnest,qin2021fcanet} are proposed to optimize spatial attention and channel attention. Most of the above works regard attention mechanism as an additional module, and with the development of the transformer \cite{vaswani2017attention}, a large number of works \cite{yu2022metaformer,dosovitskiy2020image,touvron2022resmlp,huang2023understanding,zhong2023spem,zhong2023esa} regard the attention as important parts of the backbone network.

\noindent{\textbf{Structural re-parameterization}} enables different architectures to be mutually converted through the equivalent transformation of parameters \cite{hu2022online}. For instance, a branch of 1$\times$1 convolution and a branch of 3$\times$3 convolution can be transferred into a single branch of 3×3 convolution \cite{ding2021repvgg}. 
In the training phase, multi-branch \cite{ding2021diverse,ding2021repvgg} and multi-layer \cite{Guo20, Cao20} topologies are designed to replace the vanilla layers for augmenting models. Afterward, during inference, the training-time complex models are transferred to simple ones for faster inference. Cao et al \cite{Cao20} have discussed how to merge a depthwise separable convolution kernel during training. 
Thanks to the efficiency of structural re-parameterization, it has gained great importance and has been utilized in various tasks \cite{huang2022dyrep,luo2022few,zhou2022cmb,zhang2022cs} such as compact model design \cite{dosovitskiy2020image}, architecture search \cite{Chen19, Zhang21,huang2022lottery}, pruning \cite{Ding21resrep,he2021blending,huang2021rethinking}, image recognition \cite{ding2021repmlp}, and super-resolution \cite{wang2022repsr,gao2023rcbsr}.

\section{Proposed method}
\label{sec:method}

Our attention-alike structural re-parameterization (ASR) is a simple-yet-effective method without introducing any additional computational cost and parameters in inference, which can share the effectiveness of any channel attention module by considering a learnable vector as input for the attention module. 
Fig.~\ref{fig:struc} depicts the sketch of ASR.
In the following subsections, we first introduce the preliminary channel attention modules in Section~\ref{sec:att}. We then detail discuss the training and inference phase of ASR in Section~\ref{sec:asr}.

\subsection{Preliminary: channel attention in vision}
\label{sec:att}

Given a single input $\mathbf{x} \in \mathbb{R}^{C \times H \times W}$ of channel attention modules (see Fig.~\ref{fig:struc}), we usually obtain the corresponding global information $\mathbf{u} \in \mathbb{R}^{C \times 1 \times 1}$ through global average pooling (GAP) following~\cite{hu2018squeeze,huang2020dianet,zhong2022mix}, where the $c$-th element of $\mathbf{u}$ is calculated as follows:

\begin{equation}
\begin{aligned}
\mathbf{u}_c = \text{GAP} (\mathbf{x}_c) = \frac{1}{H \times W} \sum^H_{i=1} \sum^W_{j=1} \mathbf{x}_{c, i, j}.
\end{aligned}
\label{eq:gap}
\end{equation}
Then, the attention value $\mathbf{v} \in \mathbb{R}^{C \times 1 \times 1}$ is calculated and leveraged to adjust the input $\mathbf{x}$ using Eq.~(\ref{eq:oriatt}). 

\begin{equation}
\begin{aligned}
\mathbf{x}^{\prime} = \mathbf{x} \odot \mathbf{v},  \text{ where } \mathbf{v} = \textcolor[RGB]{169,38,217}{\mathbf{T}_{\theta} }(\mathbf{u}) =  \sigma (\mathcal{F}_{\textcolor[RGB]{169,38,217}{\theta}}(\mathbf{u})), 
\end{aligned}
\label{eq:oriatt}
\end{equation}
where $\odot$ denotes channel-wise multiplication, $\mathcal{F}_{\textcolor[RGB]{169,38,217}{\theta}}$ is the neural network part of attention modules with learnable parameters $\textcolor[RGB]{169,38,217}{\theta}$, and $\sigma(\cdot)$ is activation function. 

\subsection{Attention-alike structural re-parameterization}
\label{sec:asr}

\noindent{\textbf{Training phase. }}
Based on \textit{Stripe Observation}, we can directly consider a learnable vector $\psi \in \mathbb{R}^{C \times 1 \times 1}$ as the input of $\mathcal{F}_{\textcolor[RGB]{169,38,217}{\theta}}$. Without  GAP, we measure the output of attention modules by $\textcolor[RGB]{169,38,217}{\mathbf{v}_{\psi,\theta}} = \sigma (\mathcal{F}_{\textcolor[RGB]{169,38,217}{\theta}}(\mathbf{\psi})) \in \mathbb{R}^{C \times 1 \times 1}$.

Then during the training phase, $\psi$ will be simultaneously updated with other learnable parameters.

\noindent{\textbf{Inference phase. }}
After training, according to the paradigm of SRP, we consider equivalently merging the attention module into the backbone, as shown in Fig.~\ref{fig:struc} (c). Since the "attention value" $\textcolor[RGB]{169,38,217}{\mathbf{v}_{\psi,\theta}}$ is a constant vector, for the various common-used modules $\mathbf{B}_{\hat{\theta}}$ in backbone, we can seamlessly find the corresponding transformation $g$ such that

\begin{equation}
\mathbf{B}_{\hat{\theta}} \odot \textcolor[RGB]{169,38,217}{\mathbf{v}_{\psi,\theta}} = \mathbf{B}_{g[\hat{\theta},\textcolor[RGB]{169,38,217}{\psi,\theta}]},
    \label{eq:attno2}
\end{equation}
for example, for input $\mathbf{x}$, if $\mathbf{B}_{\hat{\theta}}$ is a convolutional layer $\mathcal{C}$ with kernels $\mathbf{K}$ and bias $\mathbf{b}$, then Eq.(\ref{eq:attno2}) can be rewritten as 
	\begin{equation}
	\begin{aligned}
	\mathcal{C}(\mathbf{x};\mathbf{K},\mathbf{b}) \odot \textcolor[RGB]{169,38,217}{\mathbf{v}_{\psi,\theta}} &= \mathbf{x} * \mathbf{K} \odot \textcolor[RGB]{169,38,217}{\mathbf{v}_{\psi,\theta}} + \mathbf{b} \odot \textcolor[RGB]{169,38,217}{\mathbf{v}_{\psi,\theta}}\\ 
&= \mathcal{C}(\mathbf{x};\mathbf{K}\odot \textcolor[RGB]{169,38,217}{\mathbf{v}_{\psi,\theta}},\mathbf{b}\odot \textcolor[RGB]{169,38,217}{\mathbf{v}_{\psi,\theta}}),\\  
	\end{aligned}
	\label{eq:conv_asr}
	\end{equation}
where $*$ denote convolution and $\mathbf{K}\odot \textcolor[RGB]{169,38,217}{\mathbf{v}_{\psi,\theta}}$ means that the the product of $i$-th elements of  $\textcolor[RGB]{169,38,217}{\mathbf{v}_{\psi,\theta}}$ and $i$-th kernel of $\mathbf{K}$. Since the existing SRP methods mainly merge various neural network layers into a convolutional layer, from Eq.~(\ref{eq:conv_asr}), ASR is compatible with most of these SRP methods.
Moreover, since the channel attention module is generally placed after the normalization layer, we can also take batch normalization $\mathbf{BN}$ as an example, i.e., 
	\begin{equation}
	\begin{aligned}
	\mathbf{BN}(\mathbf{x};\gamma, \beta) \odot \textcolor[RGB]{169,38,217}{\mathbf{v}_{\psi,\theta}} &= \frac{(\mathbf{x}-\mu)\odot\gamma\odot \textcolor[RGB]{169,38,217}{\mathbf{v}_{\psi,\theta}}}{\sigma} +\beta\odot \textcolor[RGB]{169,38,217}{\mathbf{v}_{\psi,\theta}}\\ 
&= \mathbf{BN}(\mathbf{x}; \gamma\odot \textcolor[RGB]{169,38,217}{\mathbf{v}_{\psi,\theta}}, \beta\odot \textcolor[RGB]{169,38,217}{\mathbf{v}_{\psi,\theta}}),\\  
	\end{aligned}
	\label{eq:bn_asr}
	\end{equation}
where $\mu, \sigma, \gamma, $ and $\beta$ are the accumulated mean, standard deviation, and learned scaling factor and bias of $\mathbf{BN}$, respectively. See the \ul{\textbf{Appendix}} for more of the equivalent transformations for other $\mathbf{B}_{\hat{\theta}}$ in proposed ASR.

\section{Experiments}
\label{sec:experiment}

\subsection{Implementation details}
In this section, we employ various backbone architectures, including vanilla ResNet~\cite{he2016deep}, VGG~\cite{simonyan2014very}, ShuffleNetV2~\cite{ma2018shufflenet}, MobileNet~\cite{howard2017mobilenets}, ViT~\cite{dosovitskiy2020image}, RepVGG~\cite{ding2021repvgg}, and ResNet-ACNet~\cite{ding2019acnet}. Additionally, we use SE~\cite{hu2018squeeze}, IE~\cite{wang2021recurrent,liang2020instance}, SRM~\cite{lee2019srm}, ECA~\cite{wang2020eca}, and SPA~\cite{guo2020spanet} as the attention modules in our experiments. 
We employ several popular datasets, namely ImageNet~\cite{russakovsky2015imagenet}, STL10~\cite{coates2011analysis},  and CIFAR10/100~\cite{krizhevsky2009learning} in our experiments. Additionally, we use COCO~\cite{lin2014microsoft} in our style transfer experiments.
All experiments are performed using at least 3 runs to ensure statistical significance. We provide detailed recipes in the Appendix.

\begin{mdframed}[backgroundcolor=gray!8]
\begin{minipage}{\linewidth}
For a given attention module $\kappa$, it's not necessary for ASR($\kappa$) to outperform standalone $\kappa$ in terms of performance, and see Section \ref{sec:diss}~(3) for details.
\end{minipage}
\end{mdframed}

\subsection{ASR for various settings}
\label{sec:settings}
We use the task of image classification to evaluate the capability of ASR applied in various visual backbones and datasets as shown in Table~\ref{tab:backbone2} and Table~\ref{tab:backbone1}. To ensure a fair and intuitive comparison, the experimental settings are the same for the same backbones and all models are trained from scratch. Besides, due to the consideration of the variety of backbone structures, it's hard to formulate the application of ASR uniformly and please see the Appendix for specific details regarding ASR applied to each backbone. 

\begin{table}[htbp]
  \centering
  \caption{Top-1 accuracy (\%) of ASR on \textbf{ImageNet-1k} for models trained without any external data. ASR ($\kappa$) denotes the ASR based on the attention module $\kappa$, "params." represents the parameter size of the model, and "Image size" refers to the size of the images input into the model. The inference speed and parameter size of models with and without ASR ($\kappa$) are equal. 
  }
    \resizebox*{\linewidth}{!}{
    \begin{tabular}{lccc|lccc}
    \toprule
    Model & \#params. & Image size & Top-1 acc. & Model & \#params. & Image size & Top-1 acc. \\
    \midrule
    ResNet34 & 22M   & $224^2$ & 73.31  &  ViT-S & 22M   & $224^2$ & 80.02 \\
    ResNet34+ASR (SE) & 22M   & $224^2$ & 74.04 ({\color{red}{$\uparrow$ 0.73}}) & ViT-S+ASR (SE) & 22M   & $224^2$ & 80.33 ({\color{red}{$\uparrow$ 0.31}})\\
    ResNet34+ASR (ECA) & 22M   & $224^2$ & 73.88 ({\color{red}{$\uparrow$ 0.57}}) & ViT-S+ASR (ECA) & 22M   & $224^2$ & 80.37 ({\color{red}{$\uparrow$ 0.35}})\\
    ResNet34+ASR (SRM) & 22M   & $224^2$ & 73.96 ({\color{red}{$\uparrow$ 0.65}}) & ViT-S+ASR (SRM) & 22M   & $224^2$ & 80.51 ({\color{red}{$\uparrow$ 0.49}})\\
    \midrule
    ResNet50 & 26M   & $224^2$ & 76.13 & ViT-B & 86M   & $224^2$ & 81.42\\
    ResNet50+ASR (SE) & 26M   & $224^2$ & 76.70 ({\color{red}{$\uparrow$ 0.57}})& ViT-B+ASR (SE) & 86M   & $224^2$ & 82.54 ({\color{red}{$\uparrow$ 1.12}})\\
    ResNet50+ASR (ECA) & 26M   & $224^2$ & 76.87 ({\color{red}{$\uparrow$ 0.74}})& ViT-B+ASR (ECA) & 86M   & $224^2$ & 82.38 ({\color{red}{$\uparrow$ 0.96}})\\
    ResNet50+ASR (SRM) & 26M   & $224^2$ & 76.55 ({\color{red}{$\uparrow$ 0.42}})& ViT-B+ASR (SRM) & 86M   & $224^2$ & 82.24 ({\color{red}{$\uparrow$ 0.82}})\\
    \midrule
    ResNet101 & 45M   & $224^2$ & 77.06 & ViT-B & 86M   & $384^2$ & 82.97\\
    ResNet101+ASR (SE) & 45M   & $224^2$ & 77.79 ({\color{red}{$\uparrow$ 0.73}})& ViT-B+ASR (SE) & 86M   & $384^2$ & 83.55 ({\color{red}{$\uparrow$ 0.58}})\\
    ResNet101+ASR (ECA) & 45M   & $224^2$ & 77.63 ({\color{red}{$\uparrow$ 0.57}})& ViT-B+ASR (ECA) & 86M   & $384^2$  & 83.86 ({\color{red}{$\uparrow$ 0.89}})\\
    ResNet101+ASR (SRM) & 45M   & $224^2$ & 78.18 ({\color{red}{$\uparrow$ 1.12}})& ViT-B+ASR (SRM) & 86M   & $384^2$ & 83.71 ({\color{red}{$\uparrow$ 0.74}})\\
    \bottomrule
    \end{tabular}%
    }

  \label{tab:backbone2}%
\end{table}%

\begin{table}[t]
  \centering
    \caption{Top-1 accuracy (\%) of ASR for various datasets and models. ASR ($\kappa$) has the same inference speed as corresponding backbone. 
    }
    \resizebox*{1.0\linewidth}{!}{
    \begin{tabular}{lccc|lccc}
    \toprule
    \multicolumn{1}{l}{Model} & \multicolumn{1}{c}{STL10} & \multicolumn{1}{c}{CIFAR100} & \multicolumn{1}{c}{CIFAR10} &\multicolumn{1}{|l}{Model} & \multicolumn{1}{c}{STL10} & \multicolumn{1}{c}{CIFAR100} & \multicolumn{1}{c}{CIFAR10} \\
    \midrule
    ResNet164     & 82.38  & 74.32  & 92.97  & ShuffleNetV2     & 80.34  & 71.12  & 91.68 \\
           ASR (SE) & 83.70  ({\color{red}{$\uparrow$ 1.32}})  & 75.36  ({\color{red}{$\uparrow$ 1.04}})  & 94.47  ({\color{red}{$\uparrow$ 1.50}}) & ASR (SE) & 81.46  ({\color{red}{$\uparrow$ 1.12}})  & 71.34  ({\color{red}{$\uparrow$ 0.22}}) & 91.68  ({\color{red}{$\uparrow$ 0.00}})\\
           ASR (IE) & 85.15  ({\color{red}{$\uparrow$ 2.77}})  & 75.58  ({\color{red}{$\uparrow$ 1.26}})  & 94.39  ({\color{red}{$\uparrow$ 1.42}}) &ASR (IE) & 82.83  ({\color{red}{$\uparrow$ 2.49}})  & 71.89  ({\color{red}{$\uparrow$ 0.77}}) & 91.90  ({\color{red}{$\uparrow$ 0.22}}) \\
           ASR (SRM) & 83.76  ({\color{red}{$\uparrow$ 1.38}}) & 75.23  ({\color{red}{$\uparrow$ 0.91}})  & 94.45  ({\color{red}{$\uparrow$ 1.48}}) & ASR (SRM) & 81.28  ({\color{red}{$\uparrow$ 0.94}}) & 71.79  ({\color{red}{$\uparrow$ 0.67}}) & 91.69  ({\color{red}{$\uparrow$ 0.01}})\\
           ASR (SPA) & 83.44  ({\color{red}{$\uparrow$ 1.06}}) & 75.12  ({\color{red}{$\uparrow$ 0.80}}) & 94.65  ({\color{red}{$\uparrow$ 1.68}}) &ASR (SPA) & 82.15  ({\color{red}{$\uparrow$ 1.81}}) & 72.05  ({\color{red}{$\uparrow$ 0.93}}) & 91.71  ({\color{red}{$\uparrow$ 0.03}}) \\
    \midrule
    VGG19     & 79.23  & 72.48  & 93.15  & MobileNet      & 80.35   & 66.87  & 90.97\\
           ASR (SE) & 79.95  ({\color{red}{$\uparrow$ 0.72}}) & 73.37  ({\color{red}{$\uparrow$ 0.89}})  & 93.98  ({\color{red}{$\uparrow$ 0.83}}) & ASR (SE) & 81.18  ({\color{red}{$\uparrow$ 0.83}})  & 68.91  ({\color{red}{$\uparrow$ 2.04}}) & 91.48  ({\color{red}{$\uparrow$ 0.51}})\\
           ASR (IE) & 80.48  ({\color{red}{$\uparrow$ 1.25}})  & 73.41  ({\color{red}{$\uparrow$ 0.93}})  & 93.55  ({\color{red}{$\uparrow$ 0.40}}) &ASR (IE) & 81.38  ({\color{red}{$\uparrow$ 1.03}})  & 69.45  ({\color{red}{$\uparrow$ 2.58}})  & 91.30  ({\color{red}{$\uparrow$ 0.33}})\\
           ASR (SRM) & 80.35  ({\color{red}{$\uparrow$ 1.12}})  & 73.33  ({\color{red}{$\uparrow$ 0.85}})  & 93.60  ({\color{red}{$\uparrow$ 0.45}}) &ASR (SRM) & 81.51  ({\color{red}{$\uparrow$ 1.16}}) & 69.04  ({\color{red}{$\uparrow$ 2.17}}) & 91.09  ({\color{red}{$\uparrow$ 0.12}})\\
           ASR (SPA) & 80.23  ({\color{red}{$\uparrow$ 1.00}}) & 73.38  ({\color{red}{$\uparrow$ 0.90}})  & 93.79  ({\color{red}{$\uparrow$ 0.64}}) &ASR (SPA) & 81.35  ({\color{red}{$\uparrow$ 1.00}}) & 68.56  ({\color{red}{$\uparrow$ 1.69}}) & 91.36  ({\color{red}{$\uparrow$ 0.39}}) \\
    \bottomrule
    \end{tabular}%
    }

  \label{tab:backbone1}%
\end{table}%

\noindent\ul{\textbf{(1) For various backbones. }}
We first plug our ASR into the widely used model ResNet~\cite{he2016deep} and the popular transformer-based model ViT~\cite{dosovitskiy2020image}, respectively. Table~\ref{tab:backbone2} shows that, on ImageNet~\cite{russakovsky2015imagenet}, models with ASR achieve better performance than baselines, showing max improvements of 1.12\%, 0.96\%, and 1.12\% in accuracy across ASR (SE), ASR (ECA), and ASR (SRM). Besides, we explore the performance of ASR on deeper models ResNet164 and lightweight models like MobileNet~\cite{howard2017mobilenets}. Table~\ref{tab:backbone1} indicates that ASR has the ability to generally improve model performance.

\begin{figure}[t]
    \centering
\includegraphics[width=0.6\linewidth]{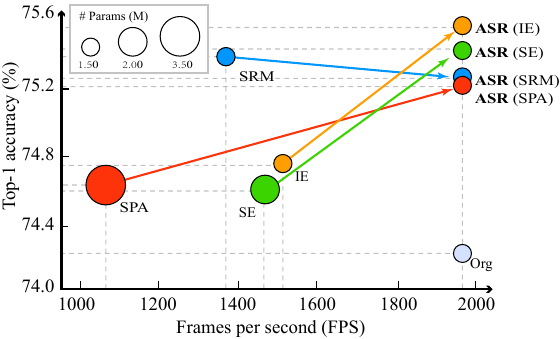}
  \caption{
  Comparison between attention modules and their ASR in the inference phase on ResNet164 and CIFAR100.
  The inference speed and \#params. of models with ASR are comparable to those of vanilla models, yet have better performance.}
\label{fig:xingneng}
\end{figure}

\noindent\ul{\textbf{(2) For various datasets. }}
We evaluate the performance of ASR across commonly used datasets to demonstrate its applicability, including ImageNet~\cite{russakovsky2015imagenet} in Table~\ref{tab:backbone2},  STL10~\cite{coates2011analysis}, CIFAR10 and CIFAR100~\cite{krizhevsky2009learning} in Table~\ref{tab:backbone1}. The results indicate that ASR is capable of improving the performance of various backbones on various datasets.

\begin{table}[htbp]
  \centering
  \caption{The experiments about the compatibility of ASR for existing attention modules. Given an attention module $\kappa$, "w/o ASR" means the backbone with only $\kappa$. "w/ ASR" denotes the backbone with both $\kappa$ and ASR ($\kappa$). $^\ddagger$ We have repeated five runs for SPA on STL10. Although its performance is relatively low compared to other modules, it still demonstrates the effectiveness of ASR.}
  \resizebox{0.99\hsize}{!}{
    \begin{tabular}{clcccc|r|clcccc}
    \toprule
          &       & \multicolumn{2}{c}{STL10} & \multicolumn{2}{c|}{CIFAR100} &       &       &       & \multicolumn{2}{c}{STL10} & \multicolumn{2}{c}{CIFAR100} \\
\cmidrule{3-6}\cmidrule{10-13}          & $\kappa$ & \multicolumn{1}{c}{w/o ASR} & \multicolumn{1}{c}{w/ ASR} & \multicolumn{1}{c}{w/o ASR} & \multicolumn{1}{c|}{w/ ASR} &       &       & $\kappa$ & \multicolumn{1}{c}{w/o ASR} & \multicolumn{1}{c}{w/ ASR} & \multicolumn{1}{c}{w/o ASR} & \multicolumn{1}{c}{w/ ASR} \\
    \midrule
    \multirow{4}[2]{*}{\begin{sideways}ResNet83\end{sideways}} & SE    & 84.21 & 85.51 ({\color{red}{$\uparrow$} 1.30}) & 74.62 & 75.87 ({\color{red}{$\uparrow$} 1.25})       &       & \multirow{4}[2]{*}{\begin{sideways}ResNet164\end{sideways}} & SE    & 83.81 & 86.41 ({\color{red}{$\uparrow$} 2.60}) & 75.29 & 77.32 ({\color{red}{$\uparrow$} 2.03})  \\
          & IE   & 84.03 & 85.70 ({\color{red}{$\uparrow$} 1.67}) & 74.74 & 75.57 ({\color{red}{$\uparrow$} 0.83})       &   &    & IE    & 84.69 & 85.99 ({\color{red}{$\uparrow$} 1.30}) & 75.78 & 76.20 ({\color{red}{$\uparrow$} 0.42})  \\
          & SRM   & 82.09 & 84.83 ({\color{red}{$\uparrow$} 2.74}) & 75.38 & 75.78 ({\color{red}{$\uparrow$} 0.40})       &       &       & SRM   & 84.85 & 85.20 ({\color{red}{$\uparrow$} 0.35}) & 75.32 & 77.18 ({\color{red}{$\uparrow$} 1.86}) \\
          & SPA   & 77.54$^\ddagger$ & 80.88 ({\color{red}{$\uparrow$} 3.34}) & 74.64 & 74.92 ({\color{red}{$\uparrow$} 0.28})       &       &       & SPA   & 75.33$^\ddagger$ & 79.61 ({\color{red}{$\uparrow$} 4.28}) & 75.48 & 77.25 ({\color{red}{$\uparrow$} 1.77})  \\
    \bottomrule
    \end{tabular}%
    }
  \label{tab:asrforatt}%
\end{table}%

\begin{figure}[t]
\centering
\includegraphics[width=0.99\linewidth]{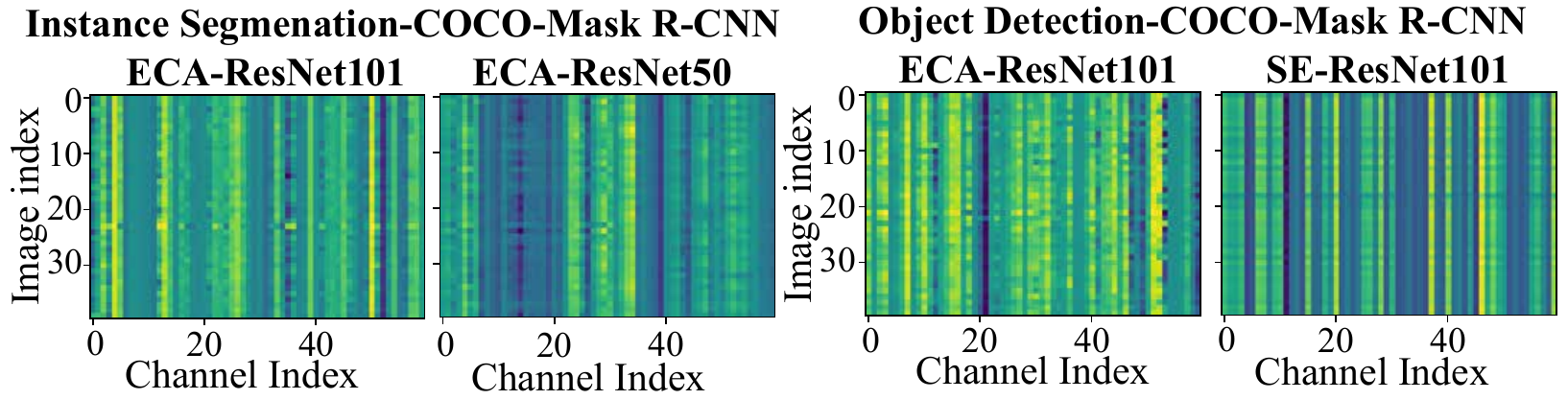}
\caption{Stripe Observation in other tasks (zoom in for best view). }
\label{fig:so}
\end{figure}

\noindent\ul{\textbf{(3) For other tasks. }}
We consider instance segmentation and object detection tasks on Mask R-CNN and COCO2017 dataset.
First, using pre-trained models from well-known third-party GitHub repositories (PytorchInsight \& ECANet), Fig.~\ref{fig:so} visualizes the \textit{Stripe Observation} of instance segmentation and object detection for first attention module under different settings, illustrating the \ul{reproducibility and generality of this phenomenon}. Moreover, Table \ref{tab:seg} further shows the effectiveness of ASR on these tasks under COCO2017 and Mask R-CNN.

\begin{table}[t]
  \centering
  \caption{The result of instance segmentation and object detection on Mask R-CNN and COCO2017. For given attention module $\kappa$, ASR($\kappa$) performs comparably to standalone $\kappa$ without any additional inference cost. It's not necessary for ASR($\kappa$) to outperform standalone $\kappa$ in terms of performance, see Section \ref{sec:diss}~(3) for details.}
  \resizebox*{1.0\linewidth}{!}{
    \begin{tabular}{lc|lc|lc|lc}
    \toprule
    \textbf{Segmentation} & \textbf{AP } & \textbf{Segmentation} & \textbf{AP} & \textbf{Detection} & \textbf{AP } & \textbf{Detection} & \textbf{AP } \\
    \midrule
    \rowcolor{Gray} ResNet50 & 34.0  & ResNet101 & 35.6  & ResNet50 & 37.3  & ResNet101 & 39.0 \\
    \midrule
    +SE    & 35.2  & +SE    & 36.8  & +SE    & 38.4  & +SE    & 40.5 \\
    +ASR (SE) & 35.1 ({\color{red}{$\uparrow$} 1.1})  & +ASR (SE) & 37.0 ({\color{red}{$\uparrow$} 1.4})  & +ASR (SE) & 38.1 ({\color{red}{$\uparrow$} 0.8}) & +ASR (SE) & 40.2 ({\color{red}{$\uparrow$} 1.2})\\
    +ASR (SE) + SE & 35.7 ({\color{red}{$\uparrow$} 1.7}) & +ASR (SE) + SE & 37.3 ({\color{red}{$\uparrow$} 1.7})  & +ASR (SE) + SE & 38.9 ({\color{red}{$\uparrow$} 1.6}) & +ASR (SE) + SE & 40.9 ({\color{red}{$\uparrow$} 1.9}) \\
    \midrule
    +ECA   & 35.4  & +ECA   & 37.1  & +ECA   & 38.8  & +ECA   & 40.7 \\
    +ASR (ECA)  & 35.6 ({\color{red}{$\uparrow$} 1.6}) & +ASR (ECA) & 36.8 ({\color{red}{$\uparrow$} 1.2}) & +ASR (ECA) & 38.6 ({\color{red}{$\uparrow$} 1.3}) & +ASR (ECA) & 40.5 ({\color{red}{$\uparrow$} 1.5})\\
    +ASR (ECA) + ECA & 35.8 ({\color{red}{$\uparrow$} 1.8}) & +ASR (ECA) + ECA & 37.4 ({\color{red}{$\uparrow$} 1.8})  & +ASR (ECA) + ECA & 39.1 ({\color{red}{$\uparrow$} 1.8}) & +ASR (ECA) + ECA & 41.1 ({\color{red}{$\uparrow$} 2.1}) \\
    \bottomrule
    \end{tabular}%
    }
  \label{tab:seg}%
\end{table}%

\noindent\ul{\textbf{(4) No burden on inference efficiency. }}
More importantly, as an attention-alike method, the performance of ASR is equivalent to or exceeds that of the original attention modules, and is consistent with the vanilla models in terms of parameter size and inference speed as shown in Fig.~\ref{fig:xingneng}. With an appropriate attention module, ASR can even outperform the attention modules in terms of accuracy for the reason that by learning constant attention that represents a learned average response, it has the potential to mitigate negative interference from the input and achieve better overall performance in an averaged sense. 

In summary, ASR relies on straightforward attention modules to enhance the performance of various models across diverse datasets without any elaborated model crafting. Furthermore, thanks to the design of structural reparameterization, ASR improves performance without any extra burden during the inference phase. Noting that we only choose simple attention modules as the training structure for SRP to showcase ASR's effectiveness in our experiments, ASR exhibits robust scalability and can be extended to incorporate diverse attention modules, having the potential for further enhancing model performance.

\subsection{The compatibility of ASR}

\begin{table}[t]
  \caption{Top-1 accuracy (\%) of ASR (SE) applied to well-known SRP methods RepVGG~\cite{ding2021repvgg} and ACNet~\cite{ding2019acnet}. RepVGGA0 and RepVGGB3 are RepVGG models with different layers of each stage and multipliers\cite{ding2021repvgg}. $^\ddagger$ We follow the settings of ResNet in the official code of ACNet, which have different layers of each stage and design (basic-block) compared to the ResNet in Table~\ref{tab:backbone1}.  }
  \centering
  \resizebox{0.5\hsize}{!}{
  
    \begin{tabular}{lcc}
    \toprule
    Model & STL10 & CIFAR100 \\
    \midrule
    RepVGGA0 & 83.48  & 68.80  \\
    RepVGGA0-ASR (SE) & 84.01  ({\color{red}{$\uparrow$ 0.53}}) & 69.38  ({\color{red}{$\uparrow$ 0.58}}) \\
    \midrule
    RepVGGB3 & 87.05  & 76.70  \\
    RepVGGB3-ASR (SE) & 88.29  ({\color{red}{$\uparrow$ 1.24}}) & 77.42 ({\color{red}{$\uparrow$ 0.72}}) \\
    \midrule
    ResNet56 $^\ddagger$ & 82.25 & 70.67 \\
    ResNet56-ACNet & 83.11  ({\color{red}{$\uparrow$ 0.86}}) & 71.42  ({\color{red}{$\uparrow$ 0.75}}) \\
    ResNet56-ACNet-ASR (SE) & 84.79  ({\color{red}{$\uparrow$ 2.54}}) & 71.81  ({\color{red}{$\uparrow$ 1.14}}) \\
    \midrule
    ResNet110 $^\ddagger$ & 82.02 & 72.60 \\
    ResNet110-ACNet & 83.54  ({\color{red}{$\uparrow$ 1.52}}) & 73.01  ({\color{red}{$\uparrow$ 0.41}}) \\
    ResNet110-ACNet-ASR (SE) & 86.13  ({\color{red}{$\uparrow$ 4.11}}) & 73.46  ({\color{red}{$\uparrow$ 0.86}}) \\
    \bottomrule
    \end{tabular}%
    
    }

  \label{tab:srp}%

\end{table}

In this section, we further investigate the compatibility of ASR with other attention modules and SRP methods.

\noindent\ul{\textbf{(1) ASR for existing attention modules. }}
As shown in Fig.~\ref{fig:struc}, ASR is an attention-like mechanism. It is natural to wonder whether ASR is compatible with DNNs that already incorporate attention modules. To investigate this, we conduct experiments as shown in Table \ref{tab:asrforatt}, and find that ASR significantly improves the performance of DNNs that have attention modules. The  SPA's low acc. on STL-10 isn't our issue (our settings are fair and repeated five run), and Table \ref{tab:asrforatt} shows even in this case ASR still works to boost SPA which further shows ASR's effectiveness.

\begin{table}[t]

  \caption{Top-1 accuracy (\%) of ResNet164 on CIFAR100. $\delta$ represents the number of ASR inserted at model's the same positions. Bold and underline indicate the best results and the second best results, respectively. }
  \centering
  \resizebox{0.8\hsize}{!}{
  
    \begin{tabular}{lcccc||ccccc}
    \toprule
    Module & $\delta=1$ &  $\delta=2$ &  $\delta=3$ &  $\delta=4$ & Module & $\delta=1$ &  $\delta=2$ &  $\delta=3$ &  $\delta=4$\\
    \midrule
    ASR (SE) & 75.36    & \textbf{75.87}    & \underline{75.72}    &  75.22 &ASR (SRM) & 75.23    & \underline{75.45}    & \textbf{75.61}    & 75.11      \\
    ASR (IE) & \underline{75.58}    & \textbf{75.71}    & 75.45    &  74.56   &ASR (SPA) & 75.12    & 75.43   & \textbf{75.81}    & \underline{75.62}   \\
    \bottomrule
    \end{tabular}%
    
    }

  \label{tab:numberofasr}%
\end{table}

\noindent\ul{\textbf{(2) ASR for existing SRP methods. }}
As a type of SRP, we also investigate the compatibility of ASR with other SRP methods. Specifically, we select RepVGG~\cite{ding2021repvgg} and ACNet~\cite{ding2019acnet}, which employ SRP in different ways and evaluate their performance with and without ASR (SE) as presented in Table~\ref{tab:srp}. The results indicate that ASR is compatible with and can further improve the performance of SRP.

In conclusion, the experiments conducted in this section highlight that ASR, as a highly scalable SRP method, possesses the ability to not only improve the performance of visual backbone networks without any extra costs during inference but also to seamlessly integrate with homogeneous methods like attention modules and other SRP techniques.

\section{Ablation study}
\label{sec:ablation}

\noindent\underline{\textbf{(1) About multiple uses of ASR. }}
Since ASR can eliminate the extra parameter and computation cost of the attention modules in the backbones during inference, theoretically, there will be no extra cost in the inference stage no matter how many times ASR is used for attention module re-parameterization. Therefore, we explore the optimal frequency denoted as $\delta$ of ASR on the model's performance. As shown in Table \ref{tab:numberofasr}, the performance of the model increases first and then decreases with the increase of $\delta$. This phenomenon suggests that ASR can further enhance the performance of the network that has already been trained with ASR, but $\delta$ cannot be too large. This is because, in general, the output of the attention module goes through a Sigmoid activation function, whose each value is not large than 1. If $\delta$ is too large, the backbone network's feature map will become small in value due to being multiplied by too many of these vectors, affecting the model's training and information forwarding. Moreover, when $\delta$ is larger, the computational cost of training increases. Therefore, we generally choose $\delta=1$ in all experiments.

\begin{table}

  \caption{ResNet164-ASR without learnable input $\psi$ on CIFAR100. $C_{i}$ and $N_{i}$ denote the initialization of unlearnable $\psi$. $C_{i}$ means that the all elements of $\psi$ are constant $i$ and $N_{i}$ means that $\psi$ samples from the normal distribution $N(\mathbf{0},i^2 \mathbf{I}_{\mathbf{dim}(\psi) })$. }
  \centering
  \resizebox{0.6\hsize}{!}{
  
    \begin{tabular}{lrrrrrrr}
    \toprule
    Module & \multicolumn{1}{l}{Ours} & \multicolumn{1}{l}{$C_{0.1}$} & \multicolumn{1}{l}{$C_{0.3}$} & \multicolumn{1}{l}{$C_{0.5}$} & \multicolumn{1}{l}{$N_{0.1}$} & \multicolumn{1}{l}{$N_{0.3}$} & \multicolumn{1}{l}{$N_{0.5}$} \\
    \midrule
    ASR (SE)    &  \textbf{75.36}     &   75.18    &    75.09   &    75.07   &   74.56    &   75.02    & 74.93 \\
    ASR (IE)    &  \textbf{75.58}     &    75.24   &    75.11   &    75.14   &   75.36    &   75.17    &  75.47 \\
    \bottomrule
    \end{tabular}%
    
    }

  \label{tab:learnable}%
\end{table}

\noindent\underline{\textbf{(2) About the learnable input $\psi$. }}
We discuss the importance of the learnable input $\psi$ for ASR. From the ASR paradigm, we verify the performance when $\psi$ is a unlearnable constant $C_{i}$ or $N_{i}$, where $C_{i}$ means that the all elements of $\psi$ are constant $i$ and $N_{i}$ means that $\psi$ samples from the normal distribution $N(\mathbf{0},i^2 \mathbf{I}_{\mathbf{dim}(\psi) })$.  The results are shown in Table \ref{tab:learnable}, where we can find the performance with constant $\psi$ is weaker relative to the learnable one, while the performance seems good when $C_{0.1}$. Therefore, in the experiments of this paper $\psi$ are initialized with $C_{0.1}$, and results about other initializations can be found in the Appendix.

\begin{figure*}[t]
  \centering
  \includegraphics[width=0.99\linewidth]{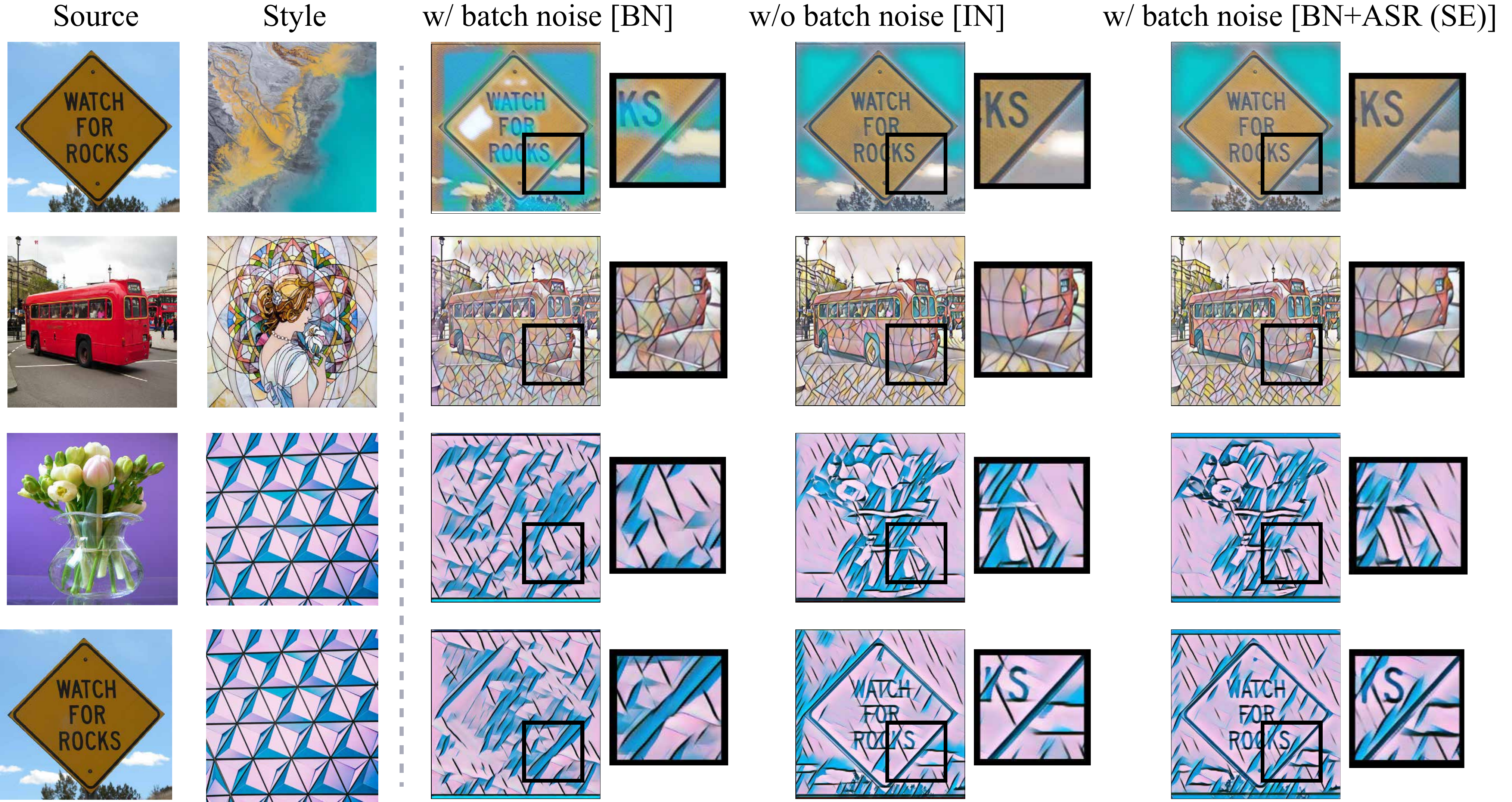}
  \caption{The results about the batch noise attack. "BN" denotes batch normalization and "IN" represents instance normalization. Batch noise severely disrupts the results of style transfer. However, even with batch noise present in BN+ASR (SE), it can yield style transfer results \ul{\textbf{comparable to IN without any batch noise}}, indicating that the proposed ASR can effectively enhance model robustness to mitigate the impact of noise. Zoom in for the best view and see Appendix for more results. }
\label{fig:fgqy}
\end{figure*}

\section{Analysis}
\label{sec:analysis}

\noindent\underline{\textbf{(1) Why the attention values approach some constant vectors?}}
In the introduction, we observe \textit{Stripe Observation} consistently across a wide range of experimental settings. This phenomenon is intriguing and although not easily explained by theory, it appears empirically reasonable. 

\begin{figure}[t]
  \centering
  \includegraphics[width=0.7\linewidth]{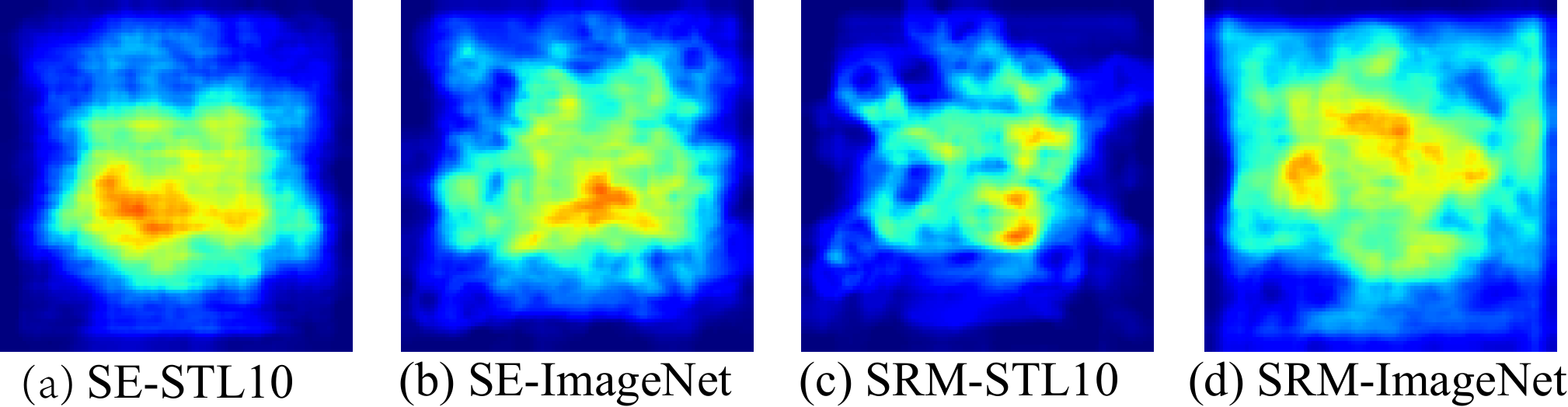}
  \caption{Average attention visualizations (using Grad-CAM~\cite{2017Grad}) on image classification task. Brighter regions indicate the areas that capture more attention from the model. }
\label{fig:meanatt}
\end{figure}

\ul{For image classification}, given a dataset $D$, certain priors such as location and color priors exist in $D$ and they have a significant impact on neural network learning, especially for attention networks that excel at capturing inductive biases. The attention mechanism adaptively adjusts the weights of feature maps to extract key information and suppress irrelevant information, which can be visualized through Grad-CAM \cite{2017Grad}. Specifically, we use the pre-trained SENet~\cite{hu2018squeeze} and SRMNet~\cite{lee2019srm} as backbones and randomly sample 500 images from the STL10 and ImageNet datasets, and measure the average attention visualizations (using Grad-CAM) of these images under the given backbones, as shown in Fig. \ref{fig:meanatt}.  We find that the regions of interest identified by the network tend to be biased toward the center of the images in all visualizations. This is because in the shooting or annotation process of these datasets, the target object $y$ was inherently present in the salient location of the image (close to the center), or else the image would not be labeled with the corresponding label of $y$. Therefore, intuitively, the existence of a series of constant "attention" values that satisfy certain priors of the network is reasonable, and these constant vectors can be regarded as the average embodiment of certain priors in $D$. 

\ul{For other tasks,} like instance segmentation and object detection, due to the nature of their tasks, their spatial priors differ significantly from image classification, meaning their labels may not necessarily be concentrated around the center of the image. However, as observed in Section \ref{sec:settings}, these tasks also exhibit the \textit{Stripe Observation}. Therefore, we believe that for these high-level tasks, the constant vectors learned by the attention module possess high-level average priors, albeit not easily visualizable. In the future, it's worth exploring the learned priors in greater detail across different tasks.

\noindent\underline{\textbf{(2) Why does ASR work? }}
ASR generates constant vectors through the attention module to help the training of DNNs, and we find that these constant vectors from ASR can regulate the noise to enhance the robustness of DNNs and help model training.

We conduct experiments on different types of noise attacks to empirically verify the ability of ASR in regulating noise to improve model robustness, including batch noise and constant noise. First, We consider the style transfer task, which generally adopts the instance normalization (IN) without batch noise, rather than BN, as adding batch noise would significantly reduce the quality of generated images due to noise interference. As shown in Fig.~\ref{fig:fgqy}, ASR can significantly alleviate the adverse effects of noise when batch noise is introduced, resulting in image quality comparable to that of IN without batch noise.

\begin{table}[t]
  \centering
      \caption{Top-1 Accuracy (\%) of vanilla ResNet164 (\textbf{Origin}) and ASR-enhanced ResNet164 on CIFAR100 under constant or random noise attack. $\sigma$ refers to the variance of the normal distribution used to generate random noise. }
    \resizebox*{0.7\linewidth}{!}{
    \begin{tabular}{ccccc}
    \toprule
    ($N_a$, $N_b$) & Origin    & ASR (SE) & ASR (IE) & ASR (SRM)  \\
    \midrule
    (1.0,0.0) & 74.32  & 75.36  ({\color{red}{$\uparrow$ \ \ 1.04}}) & 75.58  ({\color{red}{$\uparrow$ \ \ 1.26}}) & 75.23  ({\color{red}{$\uparrow$ \ \ 0.91}})  \\
    (0.8,0.8) & 45.42  & 68.81  ({\color{red}{$\uparrow$ 23.39}}) & 69.85  ({\color{red}{$\uparrow$ 24.43}}) & 69.57  ({\color{red}{$\uparrow$ 24.15}})  \\
    (0.8,0.5) & 46.10  & 71.13  ({\color{red}{$\uparrow$ 25.03}}) & 72.38  ({\color{red}{$\uparrow$ 26.28}}) & 71.69  ({\color{red}{$\uparrow$ 25.59}})  \\
    (0.5,0.5) & 35.77  & 72.18  ({\color{red}{$\uparrow$ 36.41}}) & 72.69  ({\color{red}{$\uparrow$ 36.92}}) & 71.85  ({\color{red}{$\uparrow$ 36.08}})  \\
    (0.5,0.2) & 73.10  & 74.48  ({\color{red}{$\uparrow$ \ \ 1.38}}) & 75.36  ({\color{red}{$\uparrow$ \ \ 2.26}}) & 75.02  ({\color{red}{$\uparrow$ \ \ 1.92}})  \\
    \bottomrule
    \end{tabular}%
    }

  \label{tab:noiseattack}%
\end{table}%

Next, we examine constant noise, following the settings in \cite{liang2020instance}, we inject noise $N_a$ and $N_b$ in each \textbf{BN} layer of ResNet164, i.e., $\mathbf{BN}(\mathbf{x};\gamma, \beta) = [\frac{x-\mu}{\sigma}\odot N_a + N_b]\odot\gamma + \beta$.  As shown in Table \ref{tab:noiseattack}, the noise have a large impact on neural network training. However, compared with the original network, ASR can significantly mitigate the performance loss suffered by the model.

\section{Discussion}
\label{sec:diss}

\noindent\underline{\textbf{(1) Limitation of proposed ASR.}}
ASR is inspired by the intriguing \textit{Stripe Observation}, which reveals that the channel attention values corresponding to any input in the dataset approach to some constant vectors after neural network training. However, this observation does not hold for other types of attention, like spatial attention (see Appendix). These types of attention have individual attention maps for different inputs, which means that ASR may not be directly transferable to their corresponding attention modules. Further details and comparisons can be found in the Appendix. 

\noindent\underline{\textbf{(2) It's not necessary for ASR($\kappa$) to outperform standalone $\kappa$.}}
\begin{itemize}
     \item  \ul{\textbf{ASR can use a stronger $\kappa$}}. ASR($\kappa$) has the same structure as backbone and can be paired with other strong $\kappa$ to get better performance. For instance, in Table \ref{tab:backbone1} and \ref{tab:asrforatt}, ResNet164 on CIFAR100 exhibits a standalone SRM acc. of 75.38, surpassing ASR(SRM) at 75.23. However, employing IE allows ASR(IE) with 75.58 acc. to surpass standalone SRM, while maintaining the same network structure and inference speed as Org's. 
     \item \ul{\textbf{ASR is compatible with various powerful SRP methods.}} We have shown that ASR can be compatible with  various SRP methods (Table~\ref{tab:srp}) and even itself (Table~\ref{tab:numberofasr}). Hence, ASR can be combined with various powerful SRP methods to achieve stronger performance while maintaining the same network structure as Org. For example, ACNet+ASR(SRM) can reach 75.88, surpassing the standalone SRM performance of 75.38. 

     \item \textbf{\ul{ASR is generic boosting method}}. Although ASR($\kappa$) may not awalys surpass the performance of standalone $\kappa$, it can enhance $\kappa$ such that ASR($\kappa$) + $\kappa$ surpasses standalone $\kappa$ under the same computational cost. The Table~\ref{tab:asrforatt} and Table~\ref{tab:seg} provide the details.

\end{itemize}

\section{Conclusion}
In this paper, we introduce a novel attention-alike structural re-parameterization (ASR) method, tailored for the attention mechanism, which enables effective interconversion between different network architectures. Our discovery of the \textit{Stripe Observation} provides new insights into the channel attention mechanism from a statistical perspective, leading to the development of ASR. Extensive experiments demonstrate that ASR can improve model performance\textbf{ without any extra inference cost}. We have also provided experimental evidence for the compatibility, effectiveness, and robustness of ASR, making it a promising approach for practical applications.

\clearpage
\section*{Acknowledgements}
This work was supported by National Science and Technology Major Project (2021ZD0111601), National Natural Science Foundation of China (62325605, 62272494, 623B2099), and Guangdong Basic and Applied Basic Research Foundation (2023A1515011374), and Guangzhou Science and Technology Program (2024A04J6365).


\appendix

\section{ASR for different network layers in inference phase}

In the main text, we find that the ``attention values" $\textcolor[RGB]{169,38,217}{\mathbf{v}_{\psi,\theta}}$ generated by ASR are some constant vector, for the various common-used modules $\mathbf{B}_{\hat{\theta}}$ in the backbone, we can seamlessly find the corresponding transformation $g$ such that
\begin{equation}
\mathbf{B}_{\hat{\theta}} \odot \textcolor[RGB]{169,38,217}{\mathbf{v}_{\psi,\theta}} = \mathbf{B}_{g[\hat{\theta},\textcolor[RGB]{169,38,217}{\psi,\theta}]},
\label{eq:asreq}
\end{equation}

(1) \ul{For the convolutional layer}, if $\mathbf{B}_{\hat{\theta}}$ is a convolutional layer $\mathcal{C}$ with kernels $\mathbf{K}$ and bias $\mathbf{b}$, then we have
	\begin{equation}
	\begin{aligned}
	\mathcal{C}(\mathbf{x};\mathbf{K},\mathbf{b}) \odot \textcolor[RGB]{169,38,217}{\mathbf{v}_{\psi,\theta}} & = (\mathbf{x} * \mathbf{K}) \odot \textcolor[RGB]{169,38,217}{\mathbf{v}_{\psi,\theta}} + \mathbf{b} \odot \textcolor[RGB]{169,38,217}{\mathbf{v}_{\psi,\theta}}\\
 &= \mathbf{x} * (\mathbf{K} \odot \textcolor[RGB]{169,38,217}{\mathbf{v}_{\psi,\theta}}) + \mathbf{b} \odot \textcolor[RGB]{169,38,217}{\mathbf{v}_{\psi,\theta}}\\ 
&= \mathcal{C}(\mathbf{x};\mathbf{K}\odot \textcolor[RGB]{169,38,217}{\mathbf{v}_{\psi,\theta}},\mathbf{b}\odot \textcolor[RGB]{169,38,217}{\mathbf{v}_{\psi,\theta}}),\\  
&\equiv \mathcal{C}(\mathbf{x};\mathbf{K}^\prime,\mathbf{b}^\prime)
	\end{aligned}
	\end{equation}
where $*$ denote convolution and $\mathbf{K}\odot \textcolor[RGB]{169,38,217}{\mathbf{v}_{\psi,\theta}}$ means that the the product of $i$-th elements of  $\textcolor[RGB]{169,38,217}{\mathbf{v}_{\psi,\theta}}$ and $i$-th kernel of $\mathbf{K}$. Since the existing SRP methods mainly merge various neural network layers into a convolutional layer, therefore ASR is compatible with most of these SRP methods.
\vspace{0.3cm}

(2) \ul{For the normalization layer} $\kappa$, like batch normalization \cite{ulyanov2016instance}, instance normalization \cite{ioffe2015batch}, group normalization \cite{wu2018group}, etc., they generally can be formulated as 
\begin{equation}
    \kappa(x;\mu,\sigma,\gamma,\beta) = \frac{x-\mu}{\sigma} \odot \gamma + \beta,
\end{equation}
where $\mu,\sigma,\gamma,\beta$ are the parameters of each kind of normalization method. For ASR, the Eq.(\ref{eq:asreq}) can be rewrittern as 
	\begin{equation}
	\begin{aligned}
	\kappa(\mathbf{x};\mu,\sigma,\gamma, \beta) \odot \textcolor[RGB]{169,38,217}{\mathbf{v}_{\psi,\theta}} &= \frac{(\mathbf{x}-\mu)\odot\gamma\odot \textcolor[RGB]{169,38,217}{\mathbf{v}_{\psi,\theta}}}{\sigma} +\beta\odot \textcolor[RGB]{169,38,217}{\mathbf{v}_{\psi,\theta}}\\ 
&= \kappa(\mathbf{x}; \gamma\odot \textcolor[RGB]{169,38,217}{\mathbf{v}_{\psi,\theta}}, \beta\odot \textcolor[RGB]{169,38,217}{\mathbf{v}_{\psi,\theta}}),\\  
&\equiv \kappa(\mathbf{x};\mu,\sigma,\gamma^\prime, \beta^\prime)
	\end{aligned}
	\label{eq:bn_asr}
	\end{equation}

(3) \ul{For the fully connected layer} $f(x) = Wx$, we have
	\begin{equation}
	\begin{aligned}
	f(x)\odot \textcolor[RGB]{169,38,217}{\mathbf{v}_{\psi,\theta}} &= Wx \odot \textcolor[RGB]{169,38,217}{\mathbf{v}_{\psi,\theta}}\\ 
&= (W\odot \textcolor[RGB]{169,38,217}{\mathbf{v}_{\psi,\theta}})  x \equiv W^\prime x
	\end{aligned}
	\label{eq:fcasr}
	\end{equation}

(4) \ul{For the transformer-based attention layer} $T$, we have

	\begin{align*}
	T(x;W_Q,W_K,W_V)\odot \textcolor[RGB]{169,38,217}{\mathbf{v}_{\psi,\theta}} &= \frac{W_Qx (W_Kx)^T}{\sqrt{d_k}} W_Vx\odot \textcolor[RGB]{169,38,217}{\mathbf{v}_{\psi,\theta}}\\&= \frac{W_Qx (W_Kx)^T}{\sqrt{d_k}} (W_V\odot \textcolor[RGB]{169,38,217}{\mathbf{v}_{\psi,\theta}})x\\
 &\equiv \frac{W_Qx (W_Kx)^T}{\sqrt{d_k}} W_V^\prime x \tag*{Since Eq.(\ref{eq:fcasr})}\\
 &=T(x;W_Q,W_K,W_V^\prime)
	\end{align*}

\clearpage
\section{Introdcution of implementation details}
\label{appendix:setting}

In Section~\ref{appendix:exp}, we present the experimental details, followed by an explanation in Section~\ref{appendix:apply} on how ASR is incorporated into various backbones.

\subsection{Experiment details}
\label{appendix:exp}
Unless otherwise specified, we follow the settings of~\cite{he2016deep,simonyan2014very,ding2021repvgg,ma2018shufflenet,howard2017mobilenets,ding2019acnet}.
Specifically, all models using STL10, CIFAR10, and CIFAR100 datasets with epoch set to 164. During training, we apply standard data augmentation techniques such as normalization, random cropping, and horizontal flipping. 
The batch size of CIFAR10, CIFAR100, and STL10 is 128, 128, 16, respectively. The other hyper-parameter settings of CIFAR10, CIFAR100, STL10 and ImageNet are shown in Table~\ref{tab:cifar} and Table~\ref{tab:imagenet} respectively. The patch size of ViT is 16.

\begin{table*}[ht]
    \small
    \centering
    \resizebox*{\linewidth}{!}{
    \begin{tabular}{lccccccc}
        \toprule
        & ResNet83 & ResNet164 & VGG19 & ShuffleNetV2 & MobileNet &  RepVGG & ResNet-ACNet \\
        \midrule
        optimizer & SGD (0.9) & SGD (0.9) & SGD (0.9) & SGD (0.9) & SGD (0.9)  & SGD (0.9) & SGD (0.9)\\
        schedule & 81/122 & 81/122 & 60/120/160 & 60/120/160 &  60/120/160  & 130 & cosine annealing\\
        weight decay  & 1.00E-04 & 1.00E-04 & 5.00E-04 & 5.00E-04 & 5.00E-04  & 1.00E-04 &  1.00E-04 \\
        gamma & 0.1   & 0.1   & 0.2   & 0.2   & 0.2 & 0.1 & 0.333\\
        lr     & 0.1  & 0.1   & 0.1   & 0.1   & 0.1 & 0.1  & 0.1\\
        \bottomrule                                                                                                                                                                                                                                                                                                                                                                                                                                                                                                                                                                                                                                                                                                                                                                                                                                                                                                                                                                                                                                                                                                                                                                                                                                                                                                                                                                                                                                                                                                                                                                                     
    \end{tabular}%
    }
    \caption{Implementation details for \textbf{CIFAR10/100, STL10} image classification. Normalization and standard data augmentation (random cropping and horizontal flipping) are applied to the training data. }
    \label{tab:cifar}%
\end{table*}%

\begin{table*}[ht]
    \centering
    \resizebox*{0.9\linewidth}{!}{
    \begin{tabular}{lcccccc}
        \toprule
        & ResNet34 & ResNet50 & ResNet101 & ViT-S & ViT-B & ViT-B $\uparrow$ 384 \\
        \midrule
        optimizer & SGD (0.9) & SGD (0.9) & SGD (0.9) & AdamW & AdamW & AdamW \\
        schedule & 30/60/90 & 30/60/90 & 30/60/90 & cosine annealing & cosine annealing & cosine annealing  \\
        weight decay    & 1.00E-04 & 1.00E-04 & 1.00E-04 & 5.00E-02 & 5.00E-02 & 1.00E-08 \\
        gamma & 0.1   & 0.1   & 0.1  & - & - & - \\
        lr    & 0.1   & 0.1   & 0.1 & 5.00E-04 & 5.00E-04 & 5.00E-06  \\
        epoch    & 100   & 100   & 100 & 300 & 300 & 30   \\
        batch size    & 128   & 128   & 128 & 256 & 256 & 64   \\
        \bottomrule
    \end{tabular}%
    }
    \caption{Implementation details for \textbf{ImageNet 2012} image classification. Normalization and standard data augmentation (random cropping and horizontal flipping) are applied to the training data. The random cropping of size 224 by 224 is used in these experiments. }
    \label{tab:imagenet}%
\end{table*}%

\begin{table*}[h]
    \centering
    \resizebox*{0.7\linewidth}{!}{
    \begin{tabular}{ll}
        \toprule
        Name & Explanation \\
        \midrule
        optimizer & Optimizer \\
        depth & The depth of the network \\
        schedule & Decrease learning rate at these epochs \\
        wd    & Weight decay \\
        gamma & The multiplicative factor of learning rate decay \\
        lr    & Initial learning rate \\
        \bottomrule
    \end{tabular}%
    }
    \label{tab:addlabel}%
    \caption{The additional explanation. }
\end{table*}%

\subsection{Application details}
\label{appendix:apply}

\begin{figure*}[htp]
  \centering
 \includegraphics[width=0.7\linewidth]{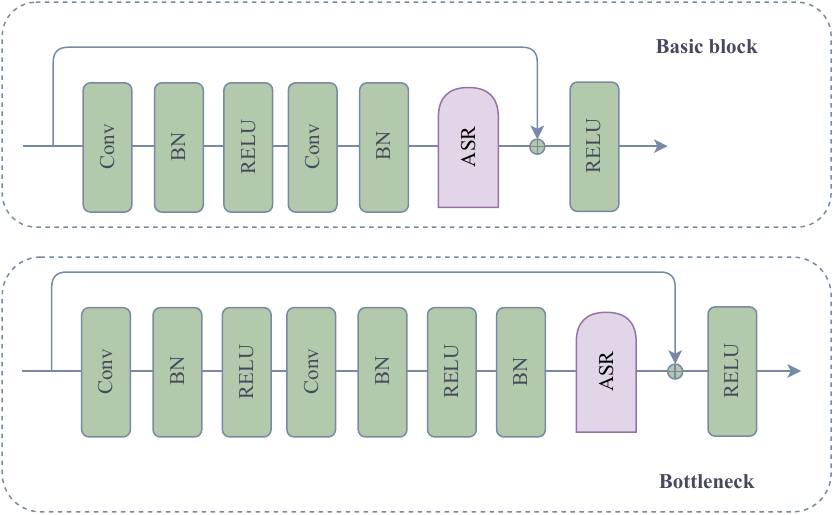}
  \caption{ASR in the blocks of ResNet during the training phase. }
\label{fig:resnet}
\end{figure*}

\begin{figure*}[htp]
  \centering
 \includegraphics[width=0.5\linewidth]{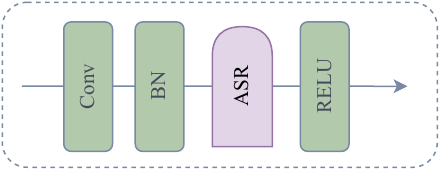}
  \caption{ASR in the block of VGG during the training phase. }
\label{fig:vgg}
\end{figure*}

\begin{figure*}[htp]
  \centering
 \includegraphics[width=0.7\linewidth]{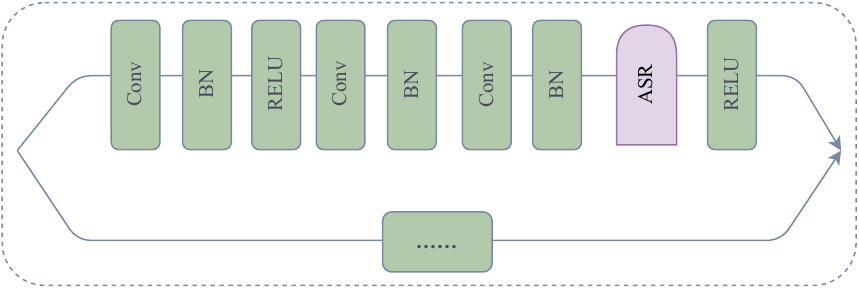}
  \caption{ASR in the block of ShuffleNetV2 during the training phase. }
\label{fig:shufflenetv2}
\end{figure*}

\begin{figure*}[htp]
  \centering
 \includegraphics[width=0.7\linewidth]{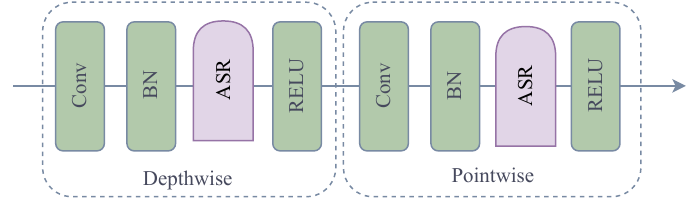}
  \caption{ASR in the block of MobileNet during the training phase. }
\label{fig:mobilenet}
\end{figure*}

\begin{figure*}[htp]
  \centering
 \includegraphics[width=0.7\linewidth]{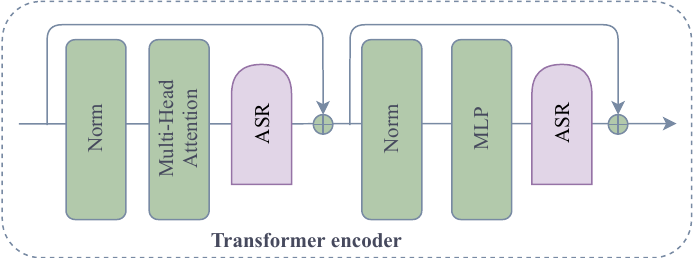}
  \caption{ASR in the transformer encoder of ViT during the training phase.  }
\label{fig:vit}
\end{figure*}

In this section, we provide detailed information on how we apply ASR to different architectures during the training phase. We focus on ResNet and ResNet-ACNet, VGG and RepVGG, ShuffleNetV2, MobileNet, and ViT. For each architecture, we specify the location of ASR within the block and provide a visualization to facilitate understanding.
Normally, the placement of ASR is typically situated post the individual layers within the model architecture. Specifically, it is conventionally positioned subsequent to batch normalization and prior to the integration of residual connections and ReLU. In essence, ASR plays the role of calibration for the outputs of each stratum within the model architecture.

\noindent{\textbf{ResNet and ResNet-ACNet }}
are popular convolutional neural network architectures that use basic blocks or bottlenecks. In our experiments, we insert ASR after the last batch normalization layer and before the residual addition operation, as shown in Fig.~\ref{fig:resnet}. We use Sigmoid as the activation function $\sigma(\cdot)$ of ASR, and we set the initial value of the learnable vector $\psi$ to 0.1.

\noindent{\textbf{VGG and RepVGG }}
are two VGG-type convolutional neural network architectures that are widely used in computer vision applications. In our experiments, we insert ASR between the batch normalization layer and ReLU activation function, as shown in Fig.~\ref{fig:vgg}. This location allows ASR to process the feature maps before they are passed to the next convolutional layer, which helps to reduce the batch noise~\cite{liang2020instance} and distortion caused by the convolution. We use Sigmoid as the activation function $\sigma(\cdot)$ of ASR, and we set the initial value of the learnable vector $\psi$ to 0.1.

\noindent{\textbf{ShuffleNetV2 }}
is a lightweight convolutional neural network architecture that uses a dual-branch structure for its block. In our experiments, we insert ASR between the last batch normalization layer and ReLU activation function in the residual branch, as shown in Fig.~\ref{fig:shufflenetv2}. We use Sigmoid as the activation function $\sigma(\cdot)$ of ASR, and we set the initial value of the learnable vector $\psi$ to 0.1.

\noindent{\textbf{MobileNet }}
is another lightweight convolutional neural network architecture that uses depthwise and pointwise convolutional layers in each block. In our experiments, we insert ASR between BN and ReLU in both layers, as demonstrated in Fig.~\ref{fig:mobilenet}. This placement allows ASR to capture the non-linearity of both convolutional layers. We use Sigmoid as the activation function $\sigma(\cdot)$ of ASR, and we set the initial value of the learnable vector $\psi$ to 0.1.

\noindent{\textbf{ViT. }}
For ViT, we apply ASR to the transformer encoder, as illustrated in Fig.~\ref{fig:vit}. ASR is inserted after multi-head attention and MLP. Moreover, distinct from other backbones, the transformer-based model ViT diverges from the conventional notion of channels. Instead, tokens and their associated features constitute the vectors from ViT. Therefore, we apply ASR to the dimension of the features. The activation function $\sigma(\cdot)$ employed within ASR is Tanh, and the initial value of the learnable vector $\psi$ is set to 1e-3. Correspondingly, the initial values of the network layers within ASR are also configured to 1e-3.

\section{The initialization of ASR}
\label{appendix:init}

In this section, we analyze the initialization of the input $\psi \in R^{C \times 1 \times 1}$ in ASR. We conduct experiments by initializing $\psi$ with values ranging from 0.1 to 0.6 and evaluate the performance of ASR on the CIFAR100 dataset as shown in Table~\ref{tab:init}. Our findings suggest that an appropriate initialization value is crucial for the performance of ASR. Specifically, we find that ASR initialized with a value of 0.1 achieves the highest accuracy of 74.83\% and 74.77\% on the test set, while the accuracy of ASR with other initialization values varies from 74.03\% to 74.73\%. These results indicate that choosing an appropriate initialization value can significantly impact the performance of ASR, and initializing ASR with a value of 0.1 leads to the best performance on the CIFAR100 dataset.

\begin{table}[htp]
  \centering
    \begin{tabular}{lcccccc}
    \toprule
    Initialization & 0.1   & 0.2   & 0.3   & 0.4   & 0.5   & 0.6 \\
    \midrule
    ASR (SE) & \textbf{74.83} & \ul{74.56} & 74.15 & 74.17 & 74.03 & 74.15 \\
    ASR (IE) & \textbf{74.77} & 74.55 & \ul{74.73} & 74.25 & 74.45 & 74.24 \\
    \bottomrule
    \end{tabular}%
  \caption{Top-1 accuracy (\%) of different initialization values on ASR's performance. The backbone is ResNet83, and the dataset is CIFAR100. Bold and underline indicate the best results and the second best results, respectively.}
  \label{tab:init}%
\end{table}%

\section{Different numbers of ASR inserted at the same position}
\label{appendix:time}

In this section, we provide additional experimental results on the impact of inserting different numbers of ASR at the same position in ResNet164 and ViT. Specifically, we evaluate the performance of ResNet164 with 1, 2, 3, and 4 ASR modules inserted at the positions shown in Fig.~\ref{fig:resnet}. We report the top-1 and top-5 accuracy on the CIFAR100 validation set.

As shown in Table~\ref{tab:time}, for ResNet164, the performance generally improves as the number of ASR modules increases when $\delta < 3$. These results suggest that inserting multiple ASR modules at the same position in ResNet164 may further enhance the performance.

\begin{table}[htbp]
  \centering
    \resizebox*{0.9\linewidth}{!}{
    \begin{tabular}{lcccccccc}
    \toprule
     \multirow{2}[3]{*}{Module} & \multicolumn{2}{c}{$\delta=1$} & \multicolumn{2}{c}{$\delta=2$} & \multicolumn{2}{c}{$\delta=3$} & \multicolumn{2}{c}{$\delta=4$} \\
\cmidrule{2-9}    & Top-1 acc. & Top-5 acc. & Top-1 acc. & Top-5 acc. & Top-1 acc. & Top-5 acc. & Top-1 acc. & Top-5 acc. \\
    \midrule
    ASR (SE) & 75.36  & 93.53  & 75.87  & 93.99  & 75.72  & 94.15  & 75.22  & 93.56  \\
    ASR (IE) & 75.58  & 93.84  & 75.71  & 93.81  & 75.45  & 94.07  & 74.56  & 93.73  \\
    ASR (SRM) & 75.23  & 93.68  & 75.45  & 94.06  & 75.61  & 94.21  & 75.11  & 93.78  \\
    ASR (SPA) & 75.12  & 93.44  & 75.43  & 93.66  & 75.81  & 94.03  & 75.62  & 93.46  \\
    \bottomrule
    \end{tabular}%
    }
  \caption{The accuracy (\%) of ResNet164 with different numbers of ASR inserted at the same position on CIFAR100. }
  \label{tab:time}%
\end{table}%

\section{Related works}

\noindent{\textbf{Attention mechanism}} selectively focuses on the most informative components of a network via self-information processing and has gained a promising performance on vision tasks \cite{liang2020instance}. 
For example, SENet \cite{hu2018squeeze} proposes the channel attention mechanism, which adjusts the feature map with channel view, and CBAM \cite{woo2018cbam} considers both channel and spatial attention for adaptive feature refinement. 
Recently, more works \cite{fu2019dual,hou2021coordinate,zhu2019empirical,gao2019global,cao2019gcnet,liang2022balancing,zhang2020resnest,qin2021fcanet,huangattns,zhong2023lsas,huang2023scalelong} are proposed to optimize spatial attention and channel attention. Most of the above works regard attention mechanism as an additional module, and with the development of the transformer \cite{vaswani2017attention}, a large number of works \cite{yu2022metaformer,dosovitskiy2020image,touvron2022resmlp,huang2023understanding,zhong2023spem,zhong2023esa} regard the attention as important parts of the backbone network.

\noindent{\textbf{Structural re-parameterization}} enables different architectures to be mutually converted through the equivalent transformation of parameters \cite{hu2022online}. For instance, a branch of 1$\times$1 convolution and a branch of 3$\times$3 convolution can be transferred into a single branch of 3×3 convolution \cite{ding2021repvgg}. 
In the training phase, multi-branch \cite{ding2021diverse,ding2021repvgg} and multi-layer \cite{Guo20, Cao20} topologies are designed to replace the vanilla layers for augmenting models. Afterward, during inference, the training-time complex models are transferred to simple ones for faster inference. Cao et al \cite{Cao20} have discussed how to merge a depthwise separable convolution kernel during training. 
Thanks to the efficiency of structural re-parameterization, it has gained great importance and has been utilized in various tasks \cite{huang2022dyrep,luo2022few,zhou2022cmb,zhang2022cs} such as compact model design \cite{dosovitskiy2020image}, architecture search \cite{Chen19, Zhang21,huang2022lottery}, pruning \cite{Ding21resrep,he2021blending,huang2021rethinking}, image recognition \cite{ding2021repmlp}, and super-resolution \cite{wang2022repsr,gao2023rcbsr}.

\section{The visualizations of the first-order difference (absolute value) for attention value over epoch }
\label{appendix:epoch}

To provide further evidence for the claim made in our paper that most of the attention values almost converge at the first learning rate decay (30 epochs), we present additional visualizations of the first-order difference (absolute value) in attention value over epoch for different structures, attention modules, datasets, and training settings (including learning rate and weight decay). Each figure includes four subplots that show the evolution of attention value for different images. The horizontal axis indicates the number of epochs, while the vertical axis represents the order of random channel ID. Unless otherwise specified, we adopt ResNet83-SE as our baseline, CIFAR100 as the default dataset, and schedule learning rate as the default learning rate, with weight decay set to 1e-4.

\begin{figure*}[ht]
  \centering
 \includegraphics[width=0.9\linewidth]{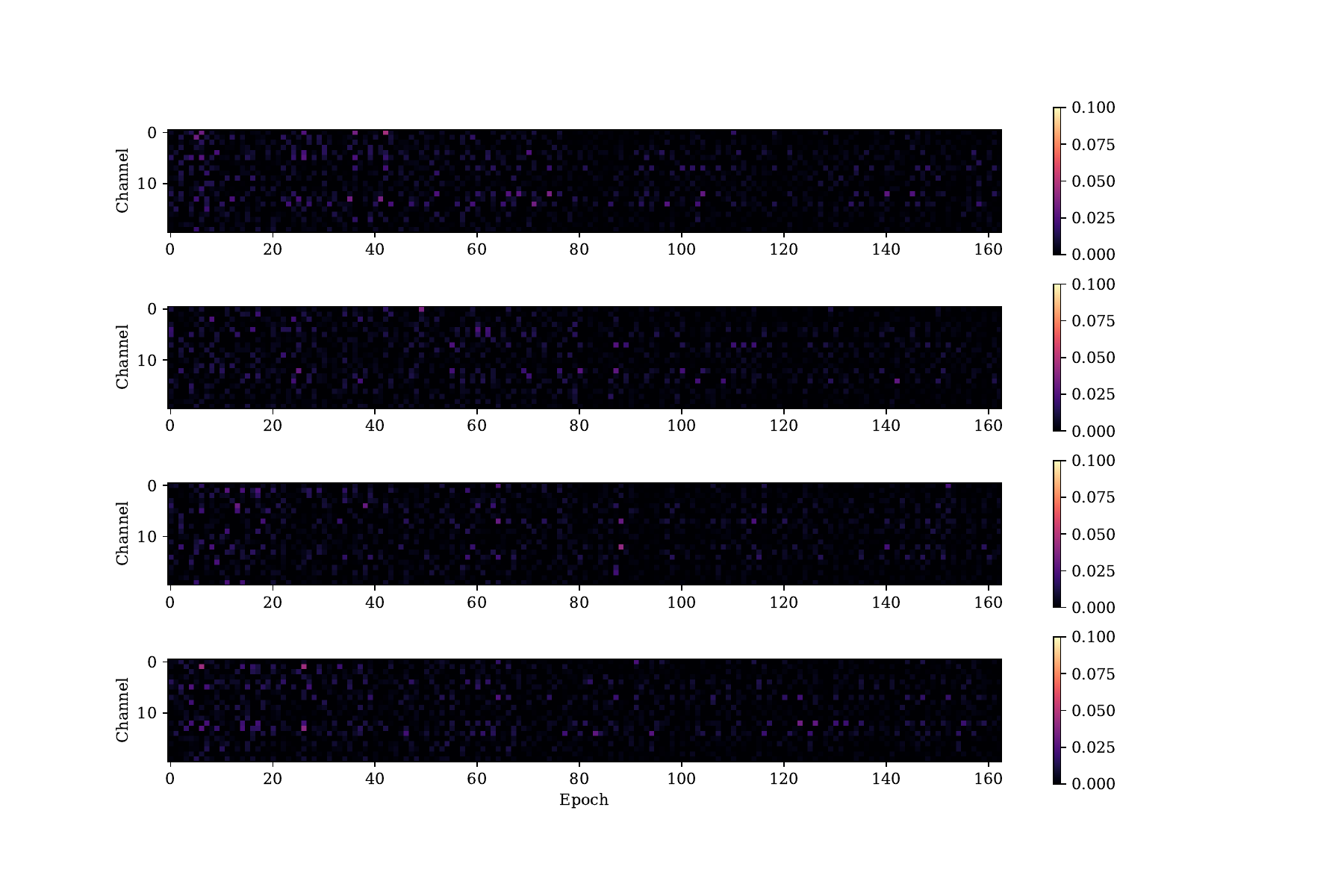}
  \caption{The visualization of the first-order difference (absolute value) for attention value of ResNet83-SE (weight decay: 1e-4) over epoch on CIFAR100. Zoom in for best view. }
\label{fig:depth83}
\end{figure*}

\begin{figure*}[ht]
  \centering
 \includegraphics[width=0.9\linewidth]{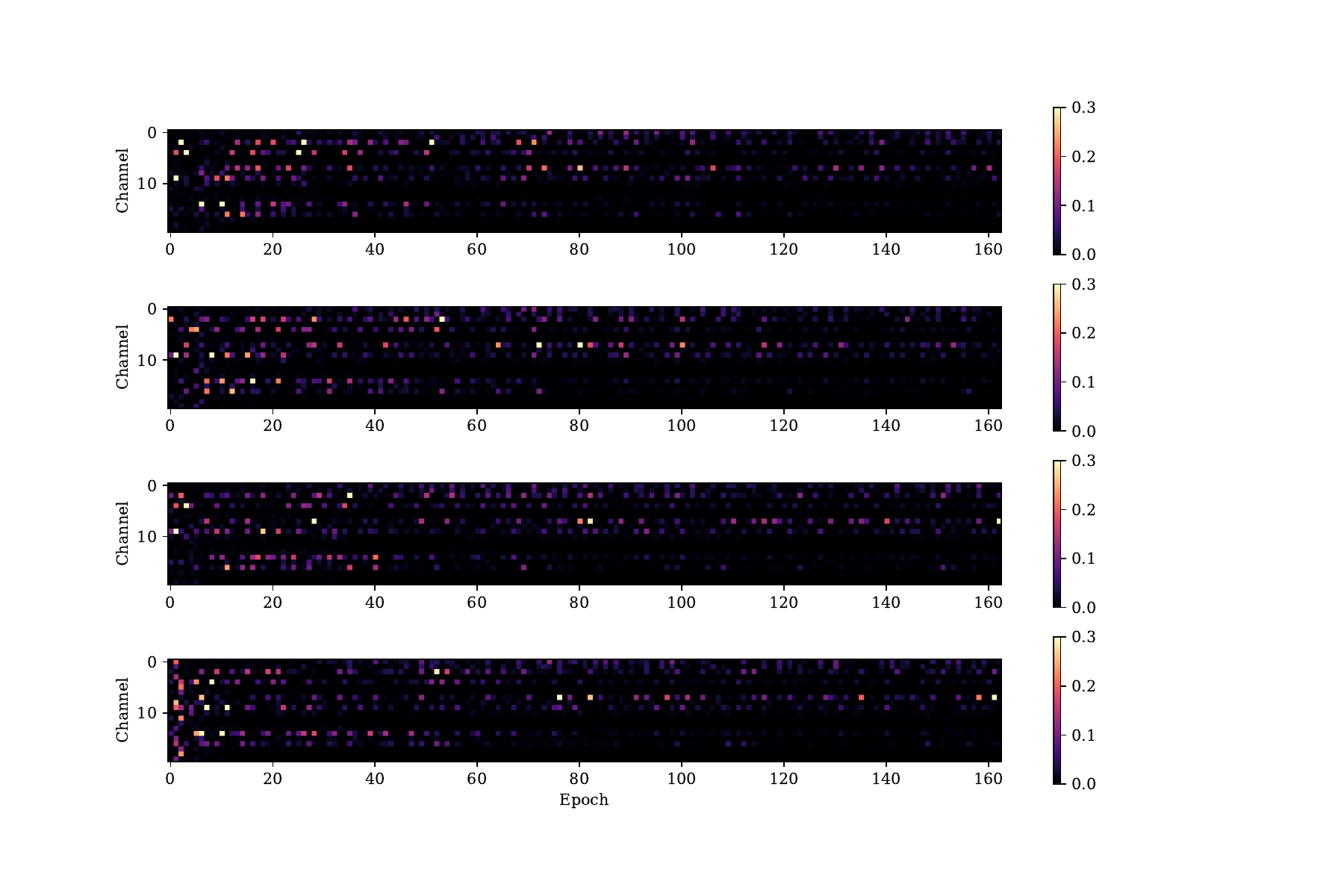}
  \caption{The visualization of the first-order difference (absolute value) for attention value of ResNet164-SE (weight decay: 1e-4) over epoch on CIFAR100. Zoom in for best view. }
\label{fig:depth164}
\end{figure*}

\noindent{\textbf{Different backbones. }}
Fig.~\ref{fig:depth83} and Fig.~\ref{fig:depth164} show the first-order difference (absolute value) in attention value over epoch for ResNet83 and ResNet164, respectively. Both backbones exhibit the same trend, indicating that most of the attention values almost converge at the first learning rate decay (30 epochs).

\begin{figure*}[ht]
  \centering
 \includegraphics[width=0.9\linewidth]{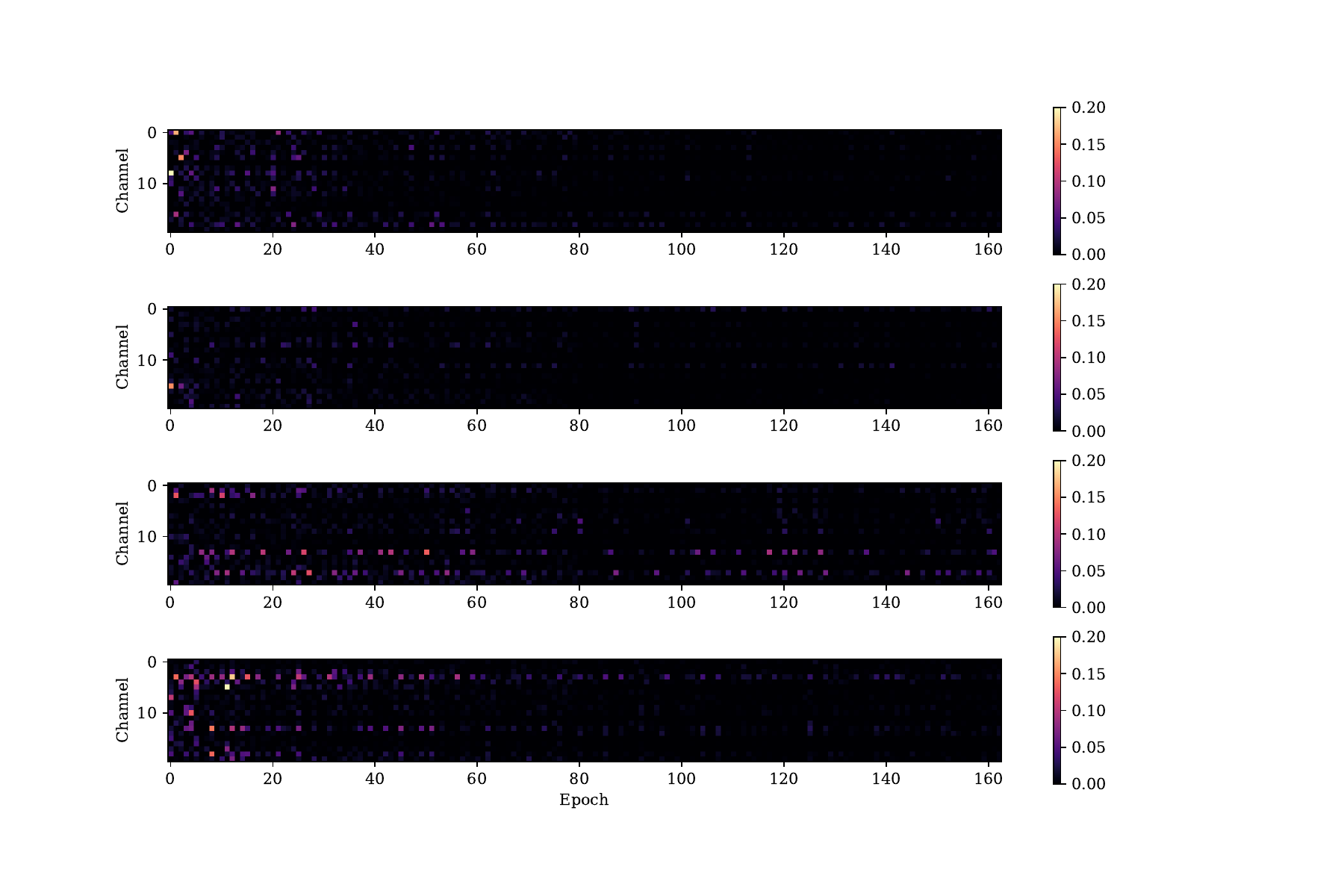}
  \caption{The visualization of the first-order difference (absolute value) for attention value of ResNet83-IE (weight decay: 1e-4) over epoch on CIFAR100. Zoom in for best view. }
\label{fig:ie-depth83}
\end{figure*}

\noindent{\textbf{Different attention modules. }}
Fig.~\ref{fig:depth83} and Fig.~\ref{fig:ie-depth83} present the first-order difference (absolute value) in attention value over epoch for two attention modules SE and IE, respectively. Although IE has some cchannel that converge more slowly than others, most of the channel attention values almost converge at 30 epochs.

\begin{figure*}[ht]
  \centering
 \includegraphics[width=0.9\linewidth]{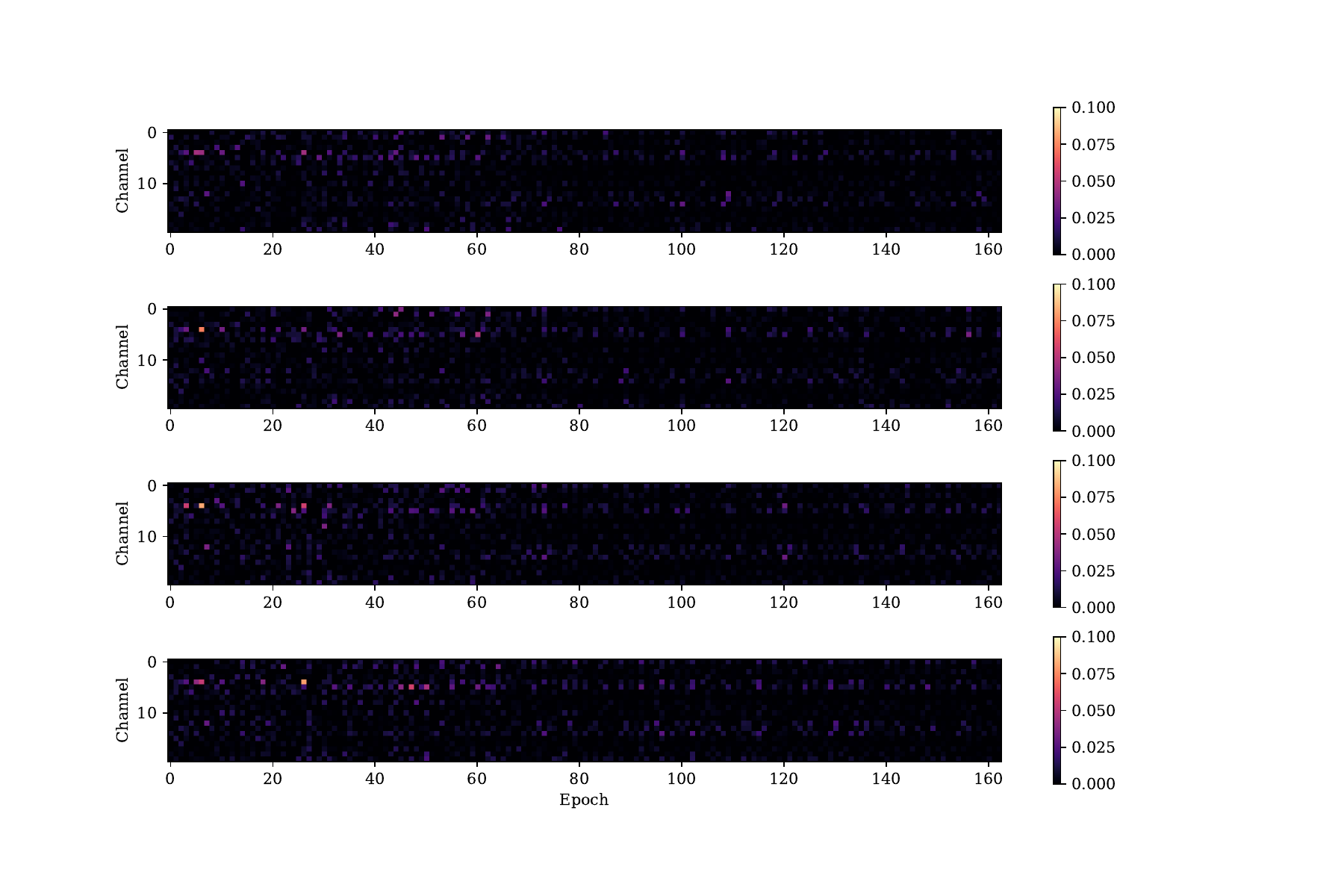}
  \caption{The visualization of the first-order difference (absolute value) for attention value of ResNet83-SE (weight decay: 1e-4) over epoch on STL10. Zoom in for best view. }
\label{fig:depth83_stl10}
\end{figure*}

\noindent{\textbf{Different datasets. }}
Fig.~\ref{fig:depth83} and Fig.~\ref{fig:depth83_stl10} compare the attention values of ResNet83-SE on CIFAR100 and STL10 during the training process. Although ResNet83-SE exhibits greater attention value fluctuations in the initial stages on the STL10, our results still align with the findings presented in our paper.

\begin{figure*}[ht]
  \centering
 \includegraphics[width=0.9\linewidth]{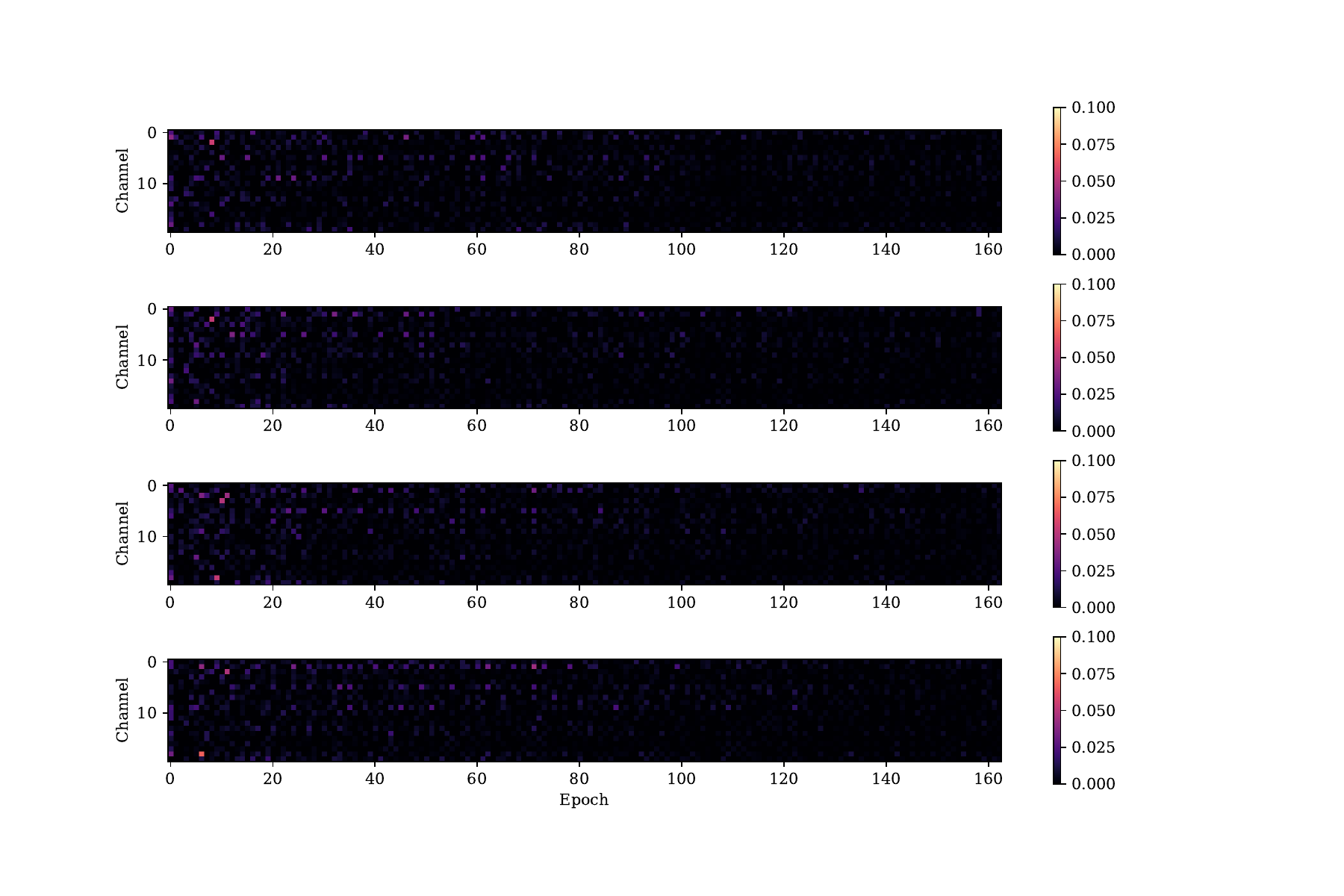}
  \caption{The visualization of the first-order difference (absolute value) for attention value of ResNet83-SE (weight decay: 1e-4) based on cosine learning rate over epoch on CIFAR100. Zoom in for best view. }
\label{fig:depth83_coslr}
\end{figure*}

\begin{figure*}[ht]
  \centering
 \includegraphics[width=0.9\linewidth]{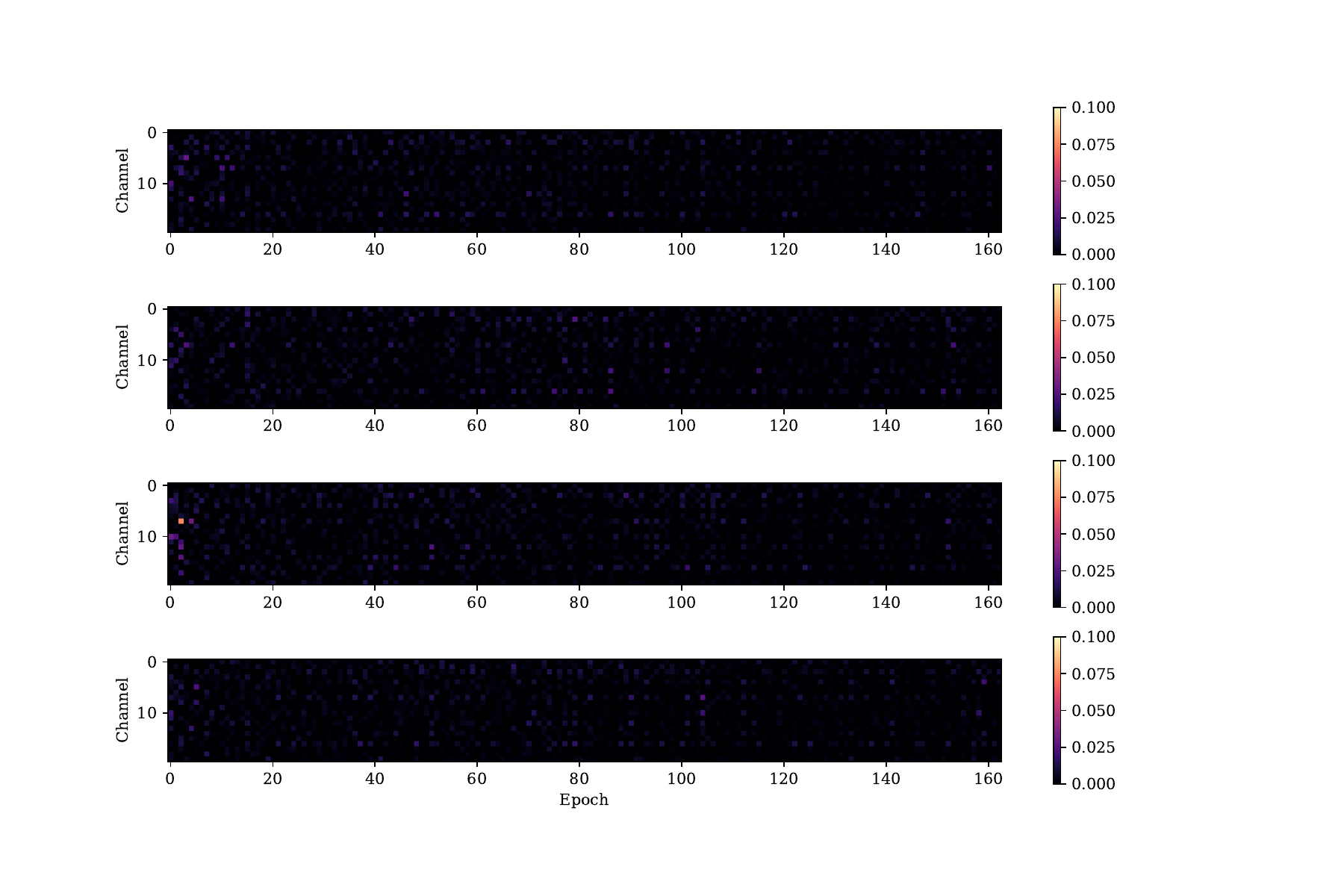}
  \caption{The visualization of the first-order difference (absolute value) for attention value of ResNet83-SE (weight decay: 2e-4) over epoch on CIFAR100. Zoom in for best view. }
\label{fig:depth83_wd2e_4}
\end{figure*}

\begin{figure*}[ht]
  \centering
 \includegraphics[width=0.9\linewidth]{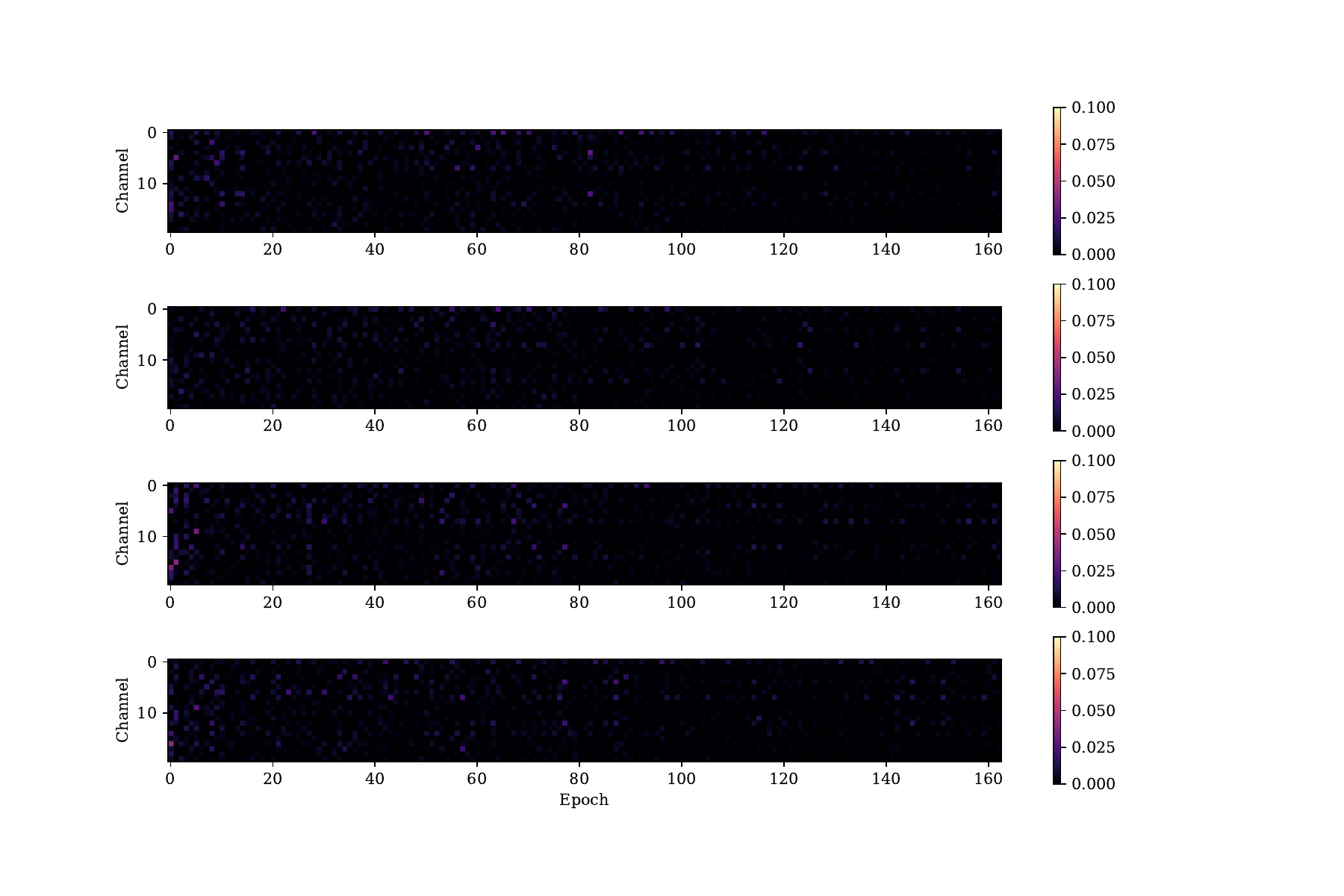}
  \caption{The visualization of the first-order difference (absolute value) for attention value of ResNet83-SE (weight decay: 3e-4) over epoch on CIFAR100. Zoom in for best view. }
\label{fig:depth83_wd3e_4}
\end{figure*}

\begin{figure*}[ht]
  \centering
 \includegraphics[width=0.9\linewidth]{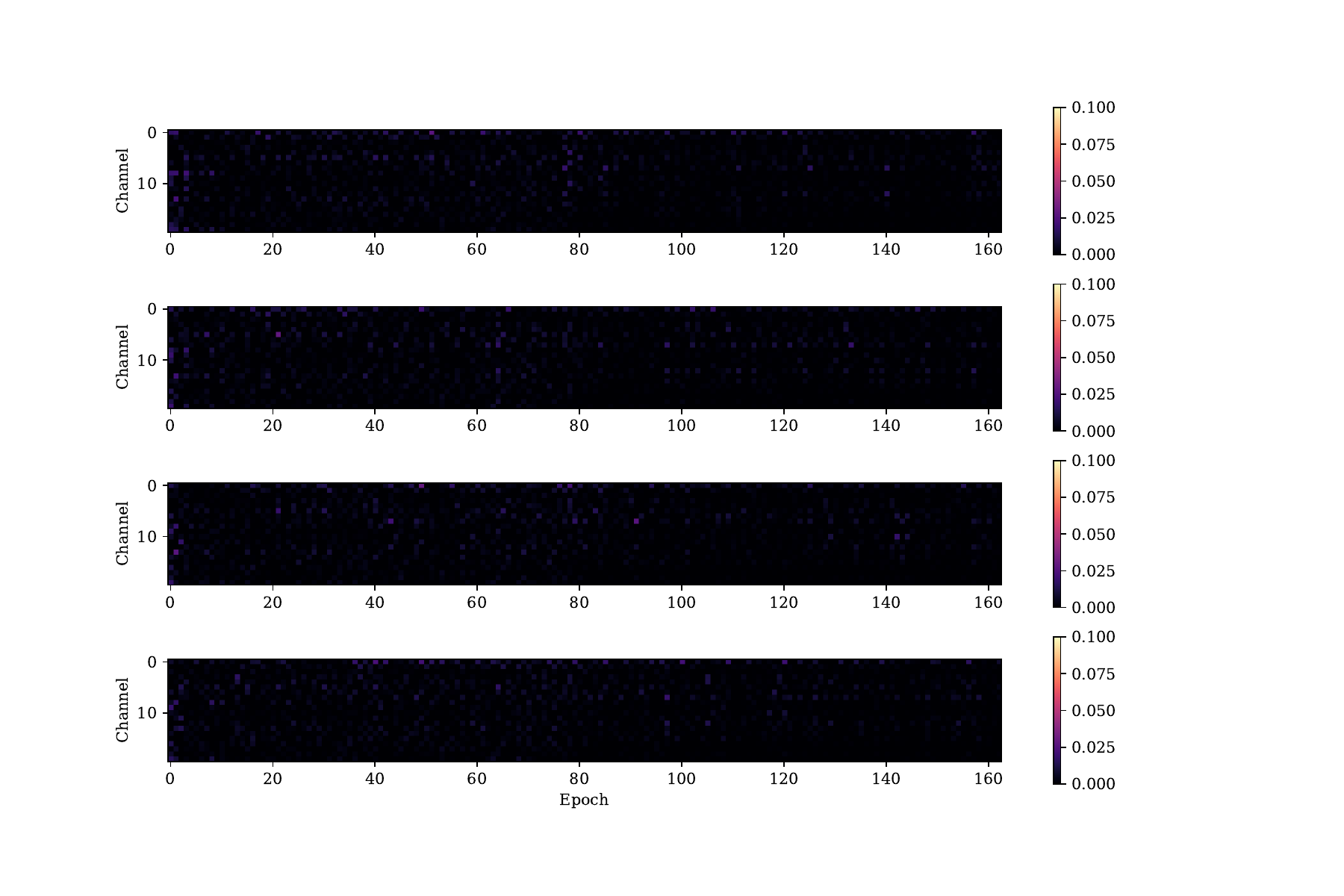}
  \caption{The visualization of the first-order difference (absolute value) for attention value of ResNet83-SE (weight decay: 4e-4) over epoch on CIFAR100. Zoom in for best view. }
\label{fig:depth83_wd4e_4}
\end{figure*}

\noindent{\textbf{Different training setting. }}
We also compare the first-order difference (absolute value) in attention value over epoch for ResNet83-SE under different training settings. Fig.~\ref{fig:depth83} and Fig.~\ref{fig:depth83_coslr} show the results of using schedule learning rate and cosine learning rate, respectively. Fig.~\ref{fig:depth83_wd2e_4}, Fig.~\ref{fig:depth83_wd3e_4}, and Fig.~\ref{fig:depth83_wd4e_4} correspond to weight decay values of 2e-4, 3e-4, and 4e-4, respectively. In all cases, we obtain results consistent with our paper's findings. We also observe that larger weight decay values lead to faster attention value convergence.

\section{The results about the batch noise attack}
\label{appendix:style}

We conduct experiments on three types of noise attacks to empirically verify the ability of ASR in regulating noise to improve model robustness, including batch noise, constant noise, and random noise. We consider the style transfer task, which generally adopts the instance normalization (IN) without batch noise, rather than BN, as adding batch noise would significantly reduce the quality of generated images due to noise interference. 

\begin{figure*}[ht]
  \centering
 \includegraphics[width=\linewidth]{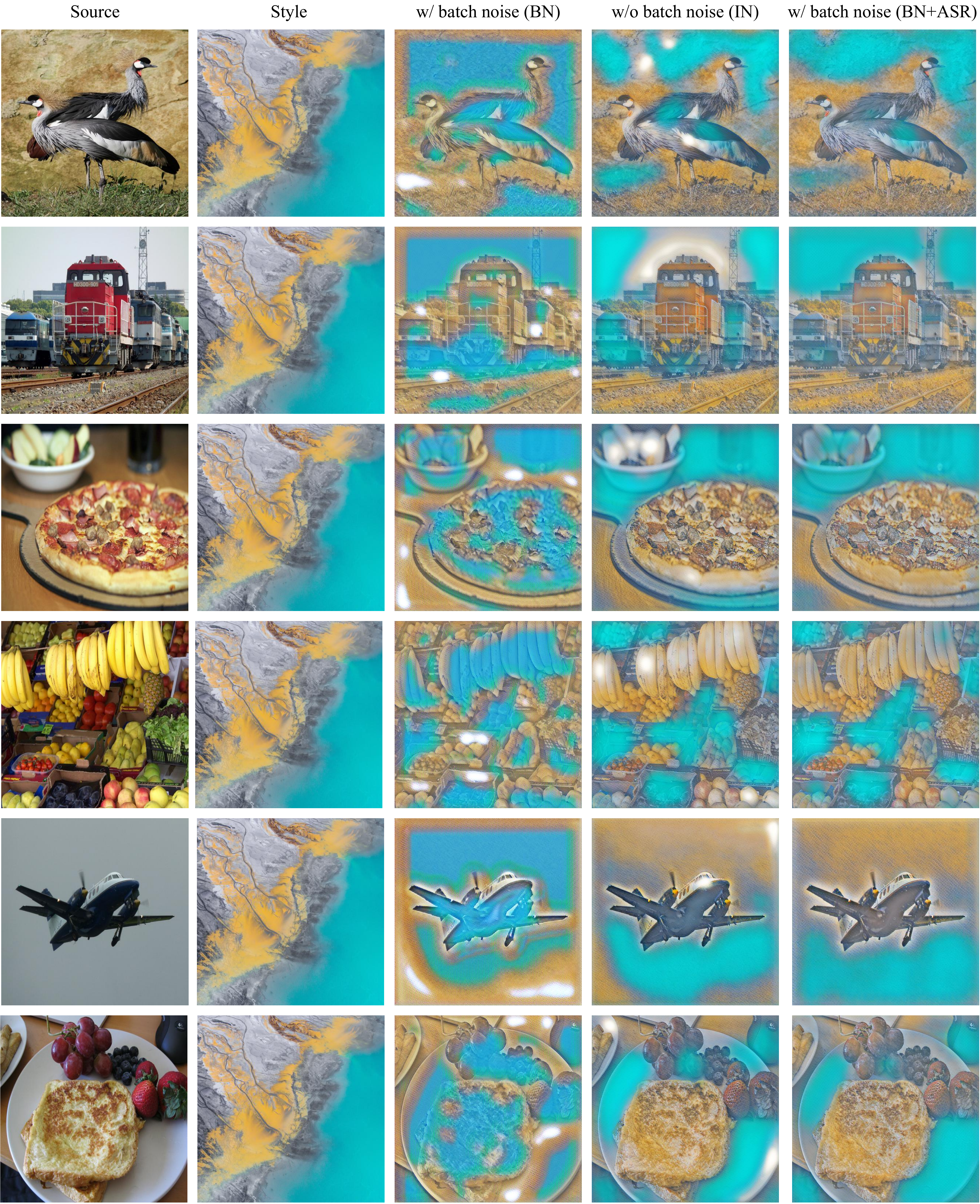}
  \caption{The results about the batch noise attack. Zoom in for best view. }
\label{fig:curioso}
\end{figure*}

\begin{figure*}[ht]
  \centering
 \includegraphics[width=\linewidth]{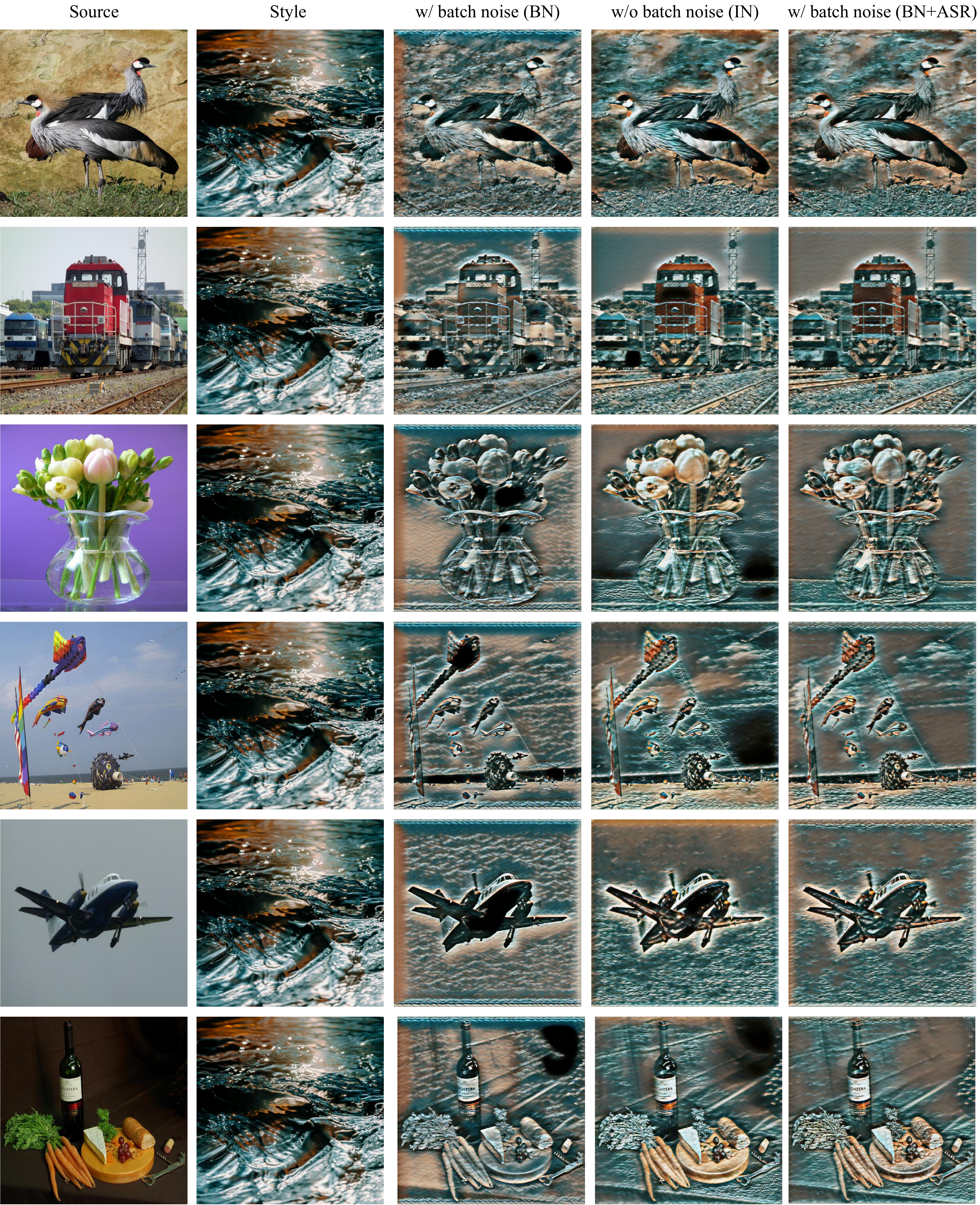}
  \caption{The results about the batch noise attack. Zoom in for best view. }
\label{fig:martinez}
\end{figure*}

\begin{figure*}[ht]
  \centering
 \includegraphics[width=\linewidth]{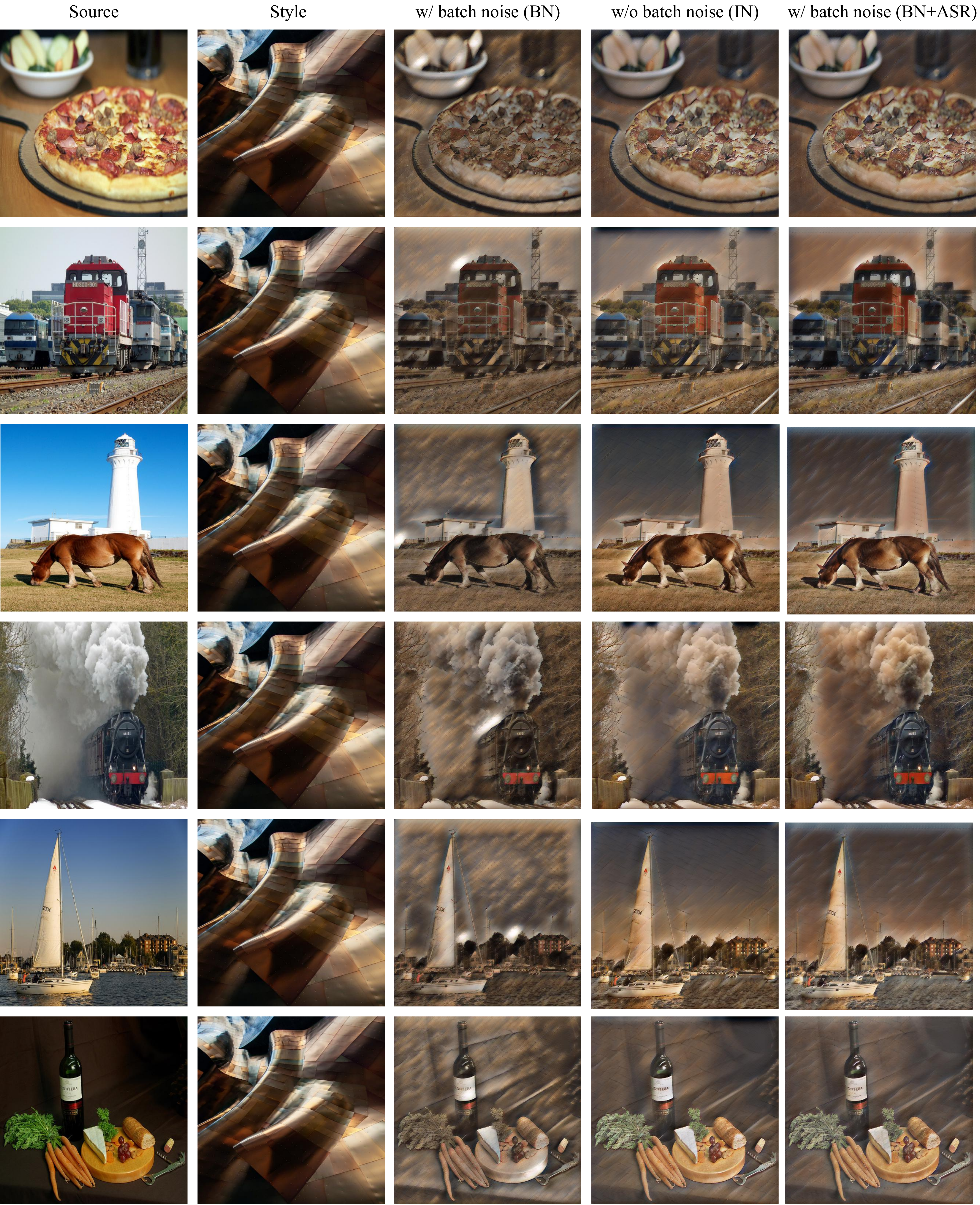}
  \caption{The results about the batch noise attack. Zoom in for best view. }
\label{fig:scott}
\end{figure*}

In this section, we present additional results on the batch noise attack to support the conclusions of our paper. As shown in Fig.~\ref{fig:curioso}, Fig.~\ref{fig:martinez}, and Fig.~\ref{fig:scott}, with batch noise (BN), there are more  blurriness compared to without batch noise (IN). However, when ASR is applied to BN, the aforementioned issues are significantly reduced. This suggests that ASR can effectively alleviate the adverse effects of noise introduced by batch normalization, resulting in image quality comparable to that of IN without batch noise.

\section{More examples of Stripe Observation}
\label{appendix:attention}

\begin{figure*}[ht]
  \centering
 \includegraphics[width=\linewidth]{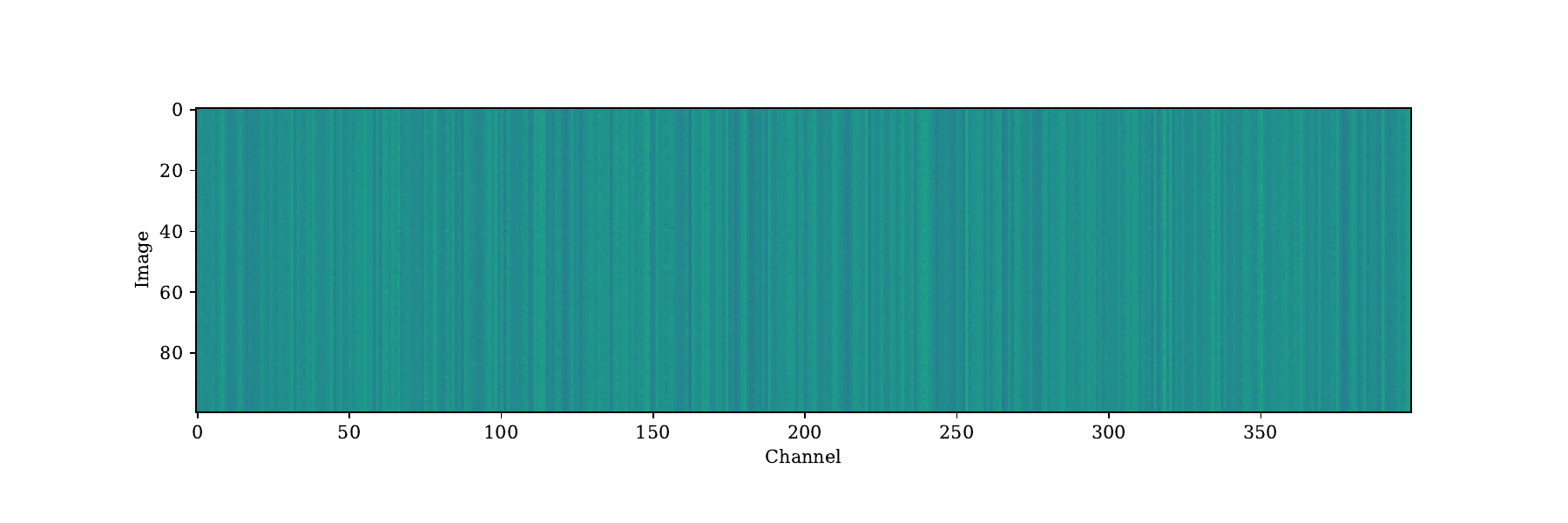}
  \caption{The attention values of different images from ResNet83-SE on CIFAR100. Zoom in for best view. }
\label{fig:stripe-senet-depth83}
\end{figure*}

\begin{figure*}[ht]
  \centering
 \includegraphics[width=\linewidth]{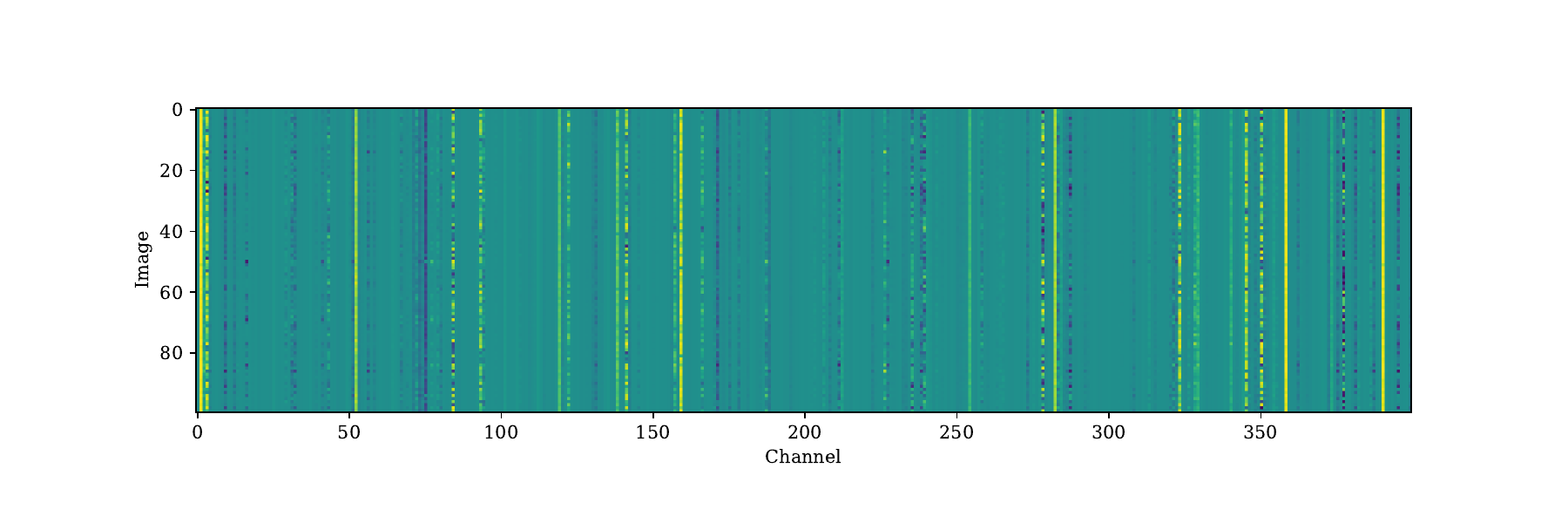}
  \caption{The attention values of different images from ResNet164-SE on CIFAR100. Zoom in for best view. }
\label{fig:stripe-senet-depth164}
\end{figure*}

\begin{figure*}[ht]
  \centering
 \includegraphics[width=\linewidth]{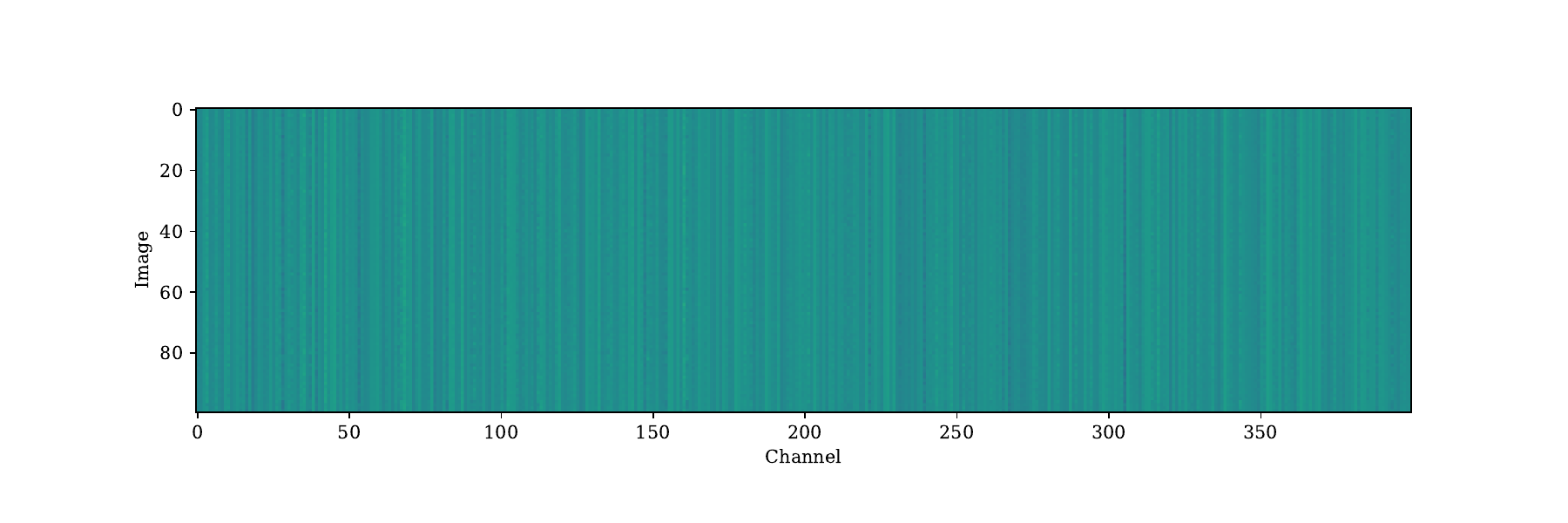}
  \caption{The attention values of different images from ResNet83-SE based on cosine learning rate on CIFAR100. Zoom in for best view. }
\label{fig:stripe-senet-depth83-coslr}
\end{figure*}

\begin{figure*}[ht]
  \centering
 \includegraphics[width=\linewidth]{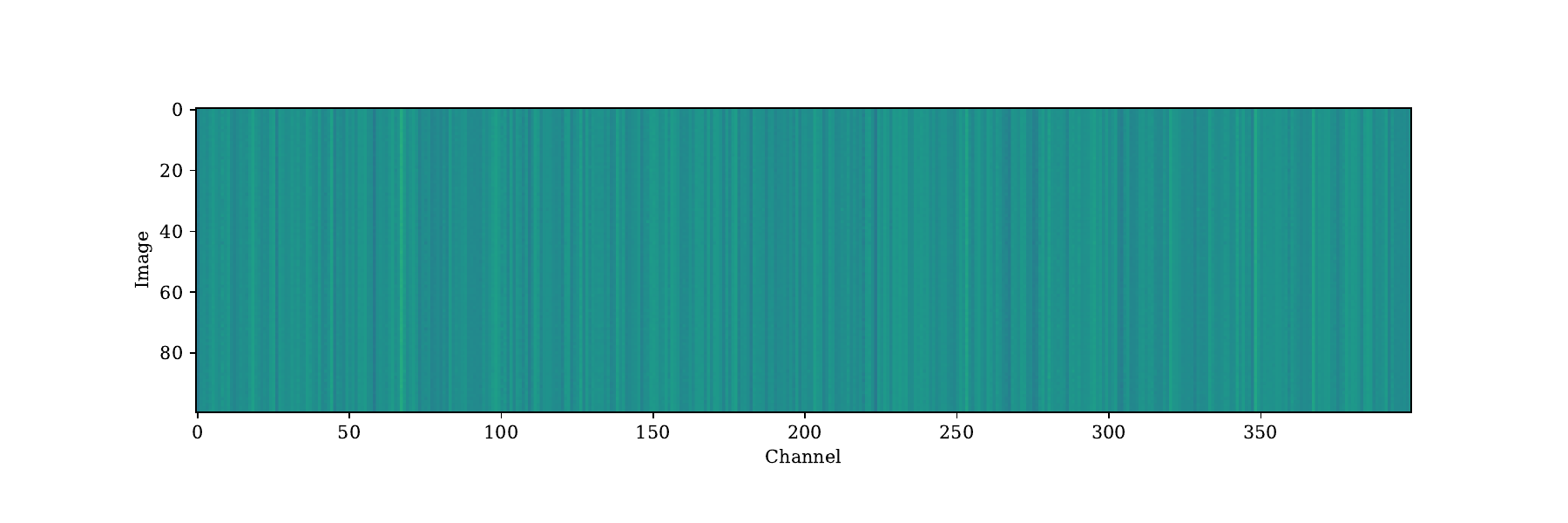}
  \caption{The attention values of different images from ResNet83-SE on STL10. Zoom in for best view. }
\label{fig:stripe-senet-depth83-stl10}
\end{figure*}

\begin{figure*}[ht]
  \centering
 \includegraphics[width=\linewidth]{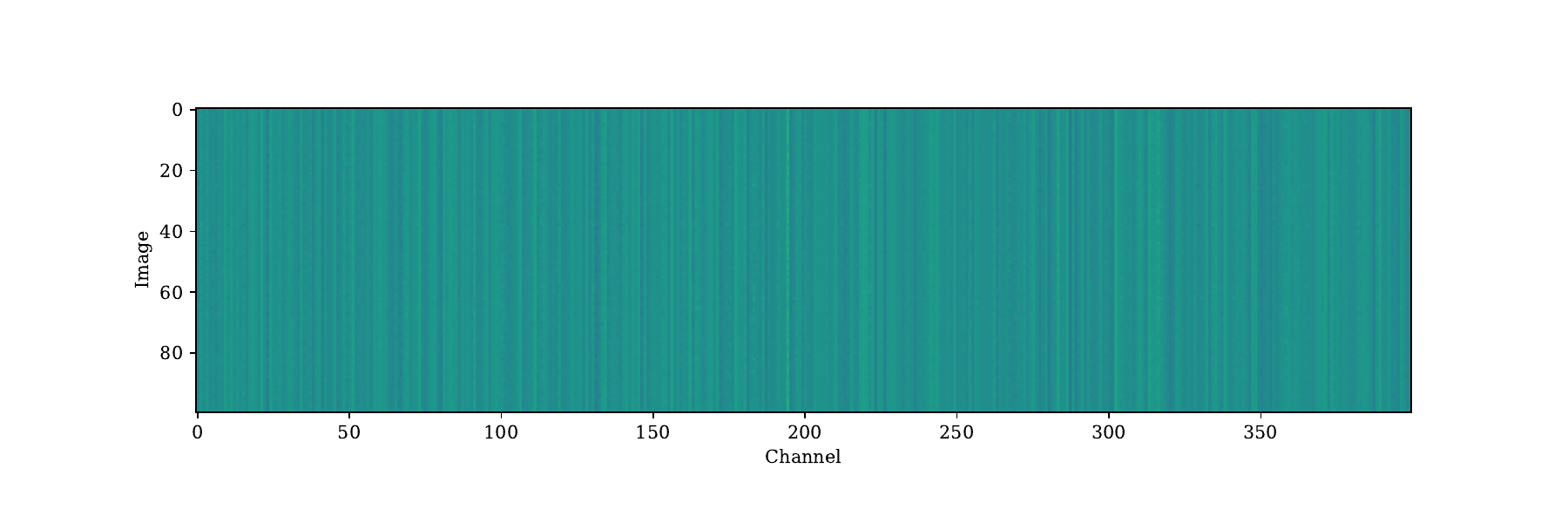}
  \caption{The attention values of different images from ResNet83-SE (weight decay: 2e-4) on CIFAR100. Zoom in for best view. }
\label{fig:stripe-senet-depth83_wd2e_4}
\end{figure*}

\begin{figure*}[ht]
  \centering
 \includegraphics[width=\linewidth]{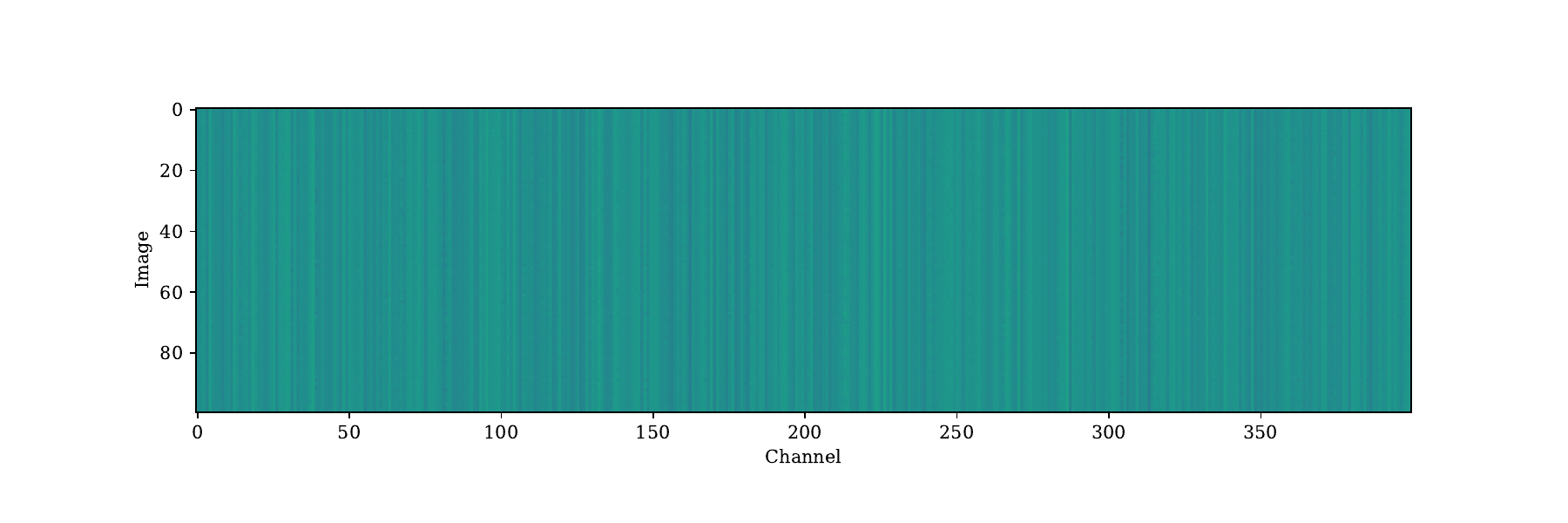}
  \caption{The attention values of different images from ResNet83-SE (weight decay: 3e-4) on CIFAR100. Zoom in for best view. }
\label{fig:stripe-senet-depth83_wd3e_4}
\end{figure*}

\begin{figure*}[ht]
  \centering
 \includegraphics[width=\linewidth]{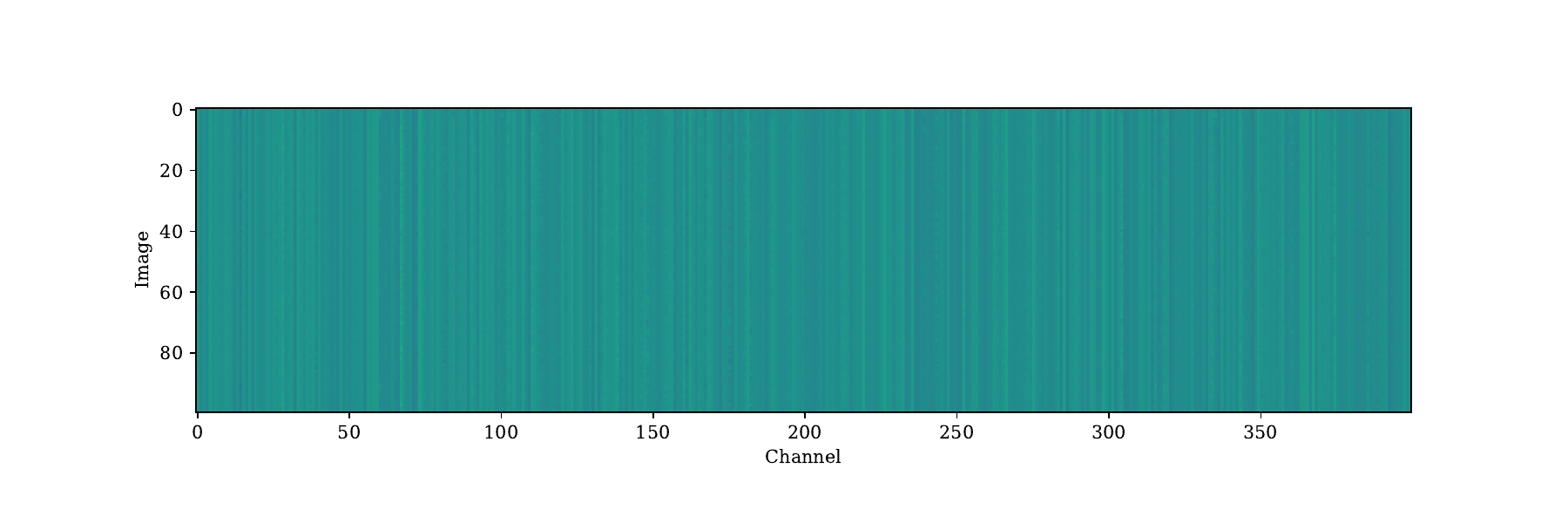}
  \caption{The attention values of different images from ResNet83-SE (weight decay: 4e-4) on CIFAR100. Zoom in for best view. }
\label{fig:stripe-senet-depth83_wd4e_4}
\end{figure*}

In this section, we present additional examples to support the conclusions of our paper that after passing through the attention module, the channel attention values of different images tend to approach a certain value within the same channel, resulting in a ``stripe structure". 
We present in Fig.~\ref{fig:stripe-senet-depth83}, Fig.~\ref{fig:stripe-senet-depth164}, Fig.~\ref{fig:stripe-senet-depth83-coslr}, Fig.~\ref{fig:stripe-senet-depth83-stl10}, Fig.~\ref{fig:stripe-senet-depth83_wd2e_4}, Fig.~\ref{fig:stripe-senet-depth83_wd3e_4}, and Fig.~\ref{fig:stripe-senet-depth83_wd4e_4} the visualization of attention values for different structures, attention modules, datasets, and training settings (including learning rate and weight decay values). The horizontal axis represents the order of random channels, while the vertical axis represents the order of random images. Corresponding to Appendix~\ref{appendix:epoch}, we use ResNet83-SE as the baseline, with the default dataset being CIFAR100, learning rate being schedule learning rate, and weight decay being 1e-4. All figures show an obvious "stripe structure," which is consistent with our conclusion in the paper that the attention values of different images tend to converge to a certain value within the same channel.

\clearpage

\begin{figure*}[ht]
  \centering
 \includegraphics[width=\linewidth]{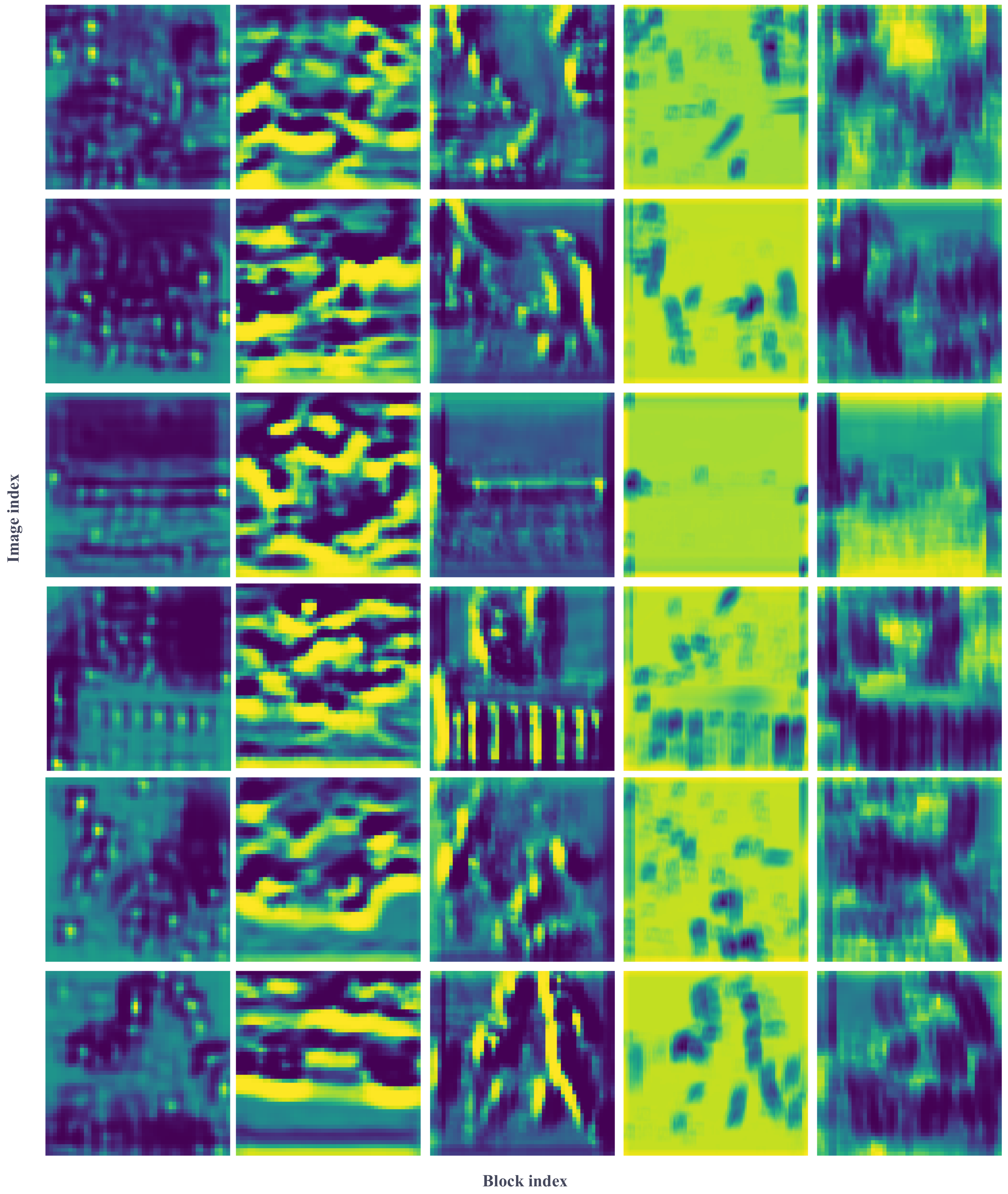}
  \caption{The visualization about the spatial attention from CBAM. We randomly select six images from STL10 and extract the spatial attention values of five blocks from ResNet83-CBAM. Each row in the visualization represents the spatial attention values of different blocks for the same image, while each column represents the spatial attention values of different images for the same block. Zoom in for best view. }
\label{fig:spatial}
\end{figure*}

\bibliographystyle{splncs04}
\bibliography{main}

\begin{thebibliography}{10}
\providecommand{\url}[1]{\texttt{#1}}
\providecommand{\urlprefix}{URL }
\providecommand{\doi}[1]{https://doi.org/#1}

\bibitem{Cao20}
Cao, J., Li, Y., Sun, M., Chen, Y., Lischinski, D., Cohen-Or, D., Chen, B., Tu, C.: Do-conv: Depthwise over-parameterized convolutional layer. arXiv 2006.12030  (2020)

\bibitem{cao2019gcnet}
Cao, Y., Xu, J., Lin, S., Wei, F., Hu, H.: Gcnet: Non-local networks meet squeeze-excitation networks and beyond. In: IEEE Conf. Comput. Vis. Worksh. pp.~0--0 (2019)

\bibitem{Chen19}
Chen, S., Chen, Y., Yan, S., Feng, J.: Efficient differentiable neural archetcture search with meta kernels. arXiv 1912.04749  (2019)

\bibitem{coates2011analysis}
Coates, A., Ng, A., Lee, H.: An analysis of single-layer networks in unsupervised feature learning. In: Proceedings of the fourteenth international conference on artificial intelligence and statistics. pp. 215--223. JMLR Workshop and Conference Proceedings (2011)

\bibitem{ding2019acnet}
Ding, X., Guo, Y., Ding, G., Han, J.: Acnet: Strengthening the kernel skeletons for powerful cnn via asymmetric convolution blocks. In: Proceedings of the IEEE/CVF international conference on computer vision. pp. 1911--1920 (2019)

\bibitem{Ding21resrep}
Ding, X., Hao, T., Tan, J., Liu, J., Han, J., Guo, Y., Ding, G.: Lossless cnn channel prunning via decoupling remembering and forgetting. In: ICCV (2021)

\bibitem{ding2021repmlp}
Ding, X., Xia, C., Zhang, X., Chu, X., Han, J., Ding, G.: Repmlp: Re-parameterizing convolutions into fully-connected layers for image recognition. arXiv preprint arXiv:2105.01883  (2021)

\bibitem{ding2021diverse}
Ding, X., Zhang, X., Han, J., Ding, G.: Diverse branch block: Building a convolution as an inception-like unit. In: Proceedings of the IEEE/CVF Conference on Computer Vision and Pattern Recognition. pp. 10886--10895 (2021)

\bibitem{ding2021repvgg}
Ding, X., Zhang, X., Ma, N., Han, J., Ding, G., Sun, J.: Repvgg: Making vgg-style convnets great again. In: Proceedings of the IEEE/CVF conference on computer vision and pattern recognition. pp. 13733--13742 (2021)

\bibitem{dosovitskiy2020image}
Dosovitskiy, A., Beyer, L., Kolesnikov, A., Weissenborn, D., Zhai, X., Unterthiner, T., Dehghani, M., Minderer, M., Heigold, G., Gelly, S., et~al.: An image is worth 16x16 words: Transformers for image recognition at scale. arXiv preprint arXiv:2010.11929  (2020)

\bibitem{fu2019dual}
Fu, J., Liu, J., Tian, H., Li, Y., Bao, Y., Fang, Z., Lu, H.: Dual attention network for scene segmentation. In: IEEE Conf. Comput. Vis. Pattern Recog. pp. 3146--3154 (2019)

\bibitem{gao2023rcbsr}
Gao, S., Zheng, C., Zhang, X., Liu, S., Wu, B., Lu, K., Zhang, D., Wang, N.: Rcbsr: re-parameterization convolution block for super-resolution. In: Computer Vision--ECCV 2022 Workshops: Tel Aviv, Israel, October 23--27, 2022, Proceedings, Part II. pp. 540--548. Springer (2023)

\bibitem{gao2019global}
Gao, Z., Xie, J., Wang, Q., Li, P.: Global second-order pooling convolutional networks. In: IEEE Conf. Comput. Vis. Pattern Recog. pp. 3024--3033 (2019)

\bibitem{guo2020spanet}
Guo, J., Ma, X., Sansom, A., McGuire, M., Kalaani, A., Chen, Q., Tang, S., Yang, Q., Fu, S.: Spanet: Spatial pyramid attention network for enhanced image recognition. In: 2020 IEEE International Conference on Multimedia and Expo (ICME). pp.~1--6. IEEE (2020)

\bibitem{Guo20}
Guo, S., Alvarze, J.M., Salzmann, M.: Expandnets: linear over re-parameterization to train compact convolutional networks. In: NeurIPS (2020)

\bibitem{he2016deep}
He, K., Zhang, X., Ren, S., Sun, J.: Deep residual learning for image recognition. In: IEEE Conf. Comput. Vis. Pattern Recog. pp. 770--778 (2016)

\bibitem{he2021blending}
He, W., Huang, Z., Liang, M., Liang, S., Yang, H.: Blending pruning criteria for convolutional neural networks. In: Artificial Neural Networks and Machine Learning--ICANN 2021: 30th International Conference on Artificial Neural Networks, Bratislava, Slovakia, September 14--17, 2021, Proceedings, Part IV 30. pp. 3--15. Springer (2021)

\bibitem{hou2021coordinate}
Hou, Q., Zhou, D., Feng, J.: Coordinate attention for efficient mobile network design. arXiv preprint arXiv:2103.02907  (2021)

\bibitem{howard2017mobilenets}
Howard, A.G., Zhu, M., Chen, B., Kalenichenko, D., Wang, W., Weyand, T., Andreetto, M., Adam, H.: Mobilenets: Efficient convolutional neural networks for mobile vision applications. arXiv preprint arXiv:1704.04861  (2017)

\bibitem{hu2018squeeze}
Hu, J., Shen, L., Sun, G.: Squeeze-and-excitation networks. In: IEEE Conf. Comput. Vis. Pattern Recog. pp. 7132--7141 (2018)

\bibitem{hu2022online}
Hu, M., Feng, J., Hua, J., Lai, B., Huang, J., Gong, X., Hua, X.S.: Online convolutional re-parameterization. In: Proceedings of the IEEE/CVF Conference on Computer Vision and Pattern Recognition. pp. 568--577 (2022)

\bibitem{huang2022dyrep}
Huang, T., You, S., Zhang, B., Du, Y., Wang, F., Qian, C., Xu, C.: Dyrep: Bootstrapping training with dynamic re-parameterization. In: Proceedings of the IEEE/CVF Conference on Computer Vision and Pattern Recognition. pp. 588--597 (2022)

\bibitem{huang2023understanding}
Huang, Z., Liang, M., Qin, J., Zhong, S., Lin, L.: Understanding self-attention mechanism via dynamical system perspective. In: Proceedings of the IEEE/CVF International Conference on Computer Vision. pp. 1412--1422 (2023)

\bibitem{huangattns}
Huang, Z., Liang, M., Zhong, S., Lin, L.: Attns: Attention-inspired numerical solving for limited data scenarios. In: Forty-first International Conference on Machine Learning (2024)

\bibitem{huang2022lottery}
Huang, Z., Liang, S., Liang, M., He, W., Yang, H., Lin, L.: The lottery ticket hypothesis for self-attention in convolutional neural network. arXiv preprint arXiv:2207.07858  (2022)

\bibitem{huang2020dianet}
Huang, Z., Liang, S., Liang, M., Yang, H.: Dianet: Dense-and-implicit attention network. In: AAAI. pp. 4206--4214 (2020)

\bibitem{huang2021rethinking}
Huang, Z., Shao, W., Wang, X., Lin, L., Luo, P.: Rethinking the pruning criteria for convolutional neural network. Advances in Neural Information Processing Systems  \textbf{34},  16305--16318 (2021)

\bibitem{huang2023scalelong}
Huang, Z., Zhou, P., Yan, S., Lin, L.: Scalelong: Towards more stable training of diffusion model via scaling network long skip connection. Advances in Neural Information Processing Systems  \textbf{36},  70376--70401 (2023)

\bibitem{ioffe2015batch}
Ioffe, S., Szegedy, C.: Batch normalization: Accelerating deep network training by reducing internal covariate shift. In: International conference on machine learning. pp. 448--456. pmlr (2015)

\bibitem{krizhevsky2009learning}
Krizhevsky, A., Hinton, G., et~al.: Learning multiple layers of features from tiny images  (2009)

\bibitem{lee2019srm}
Lee, H., Kim, H.E., Nam, H.: Srm: A style-based recalibration module for convolutional neural networks. In: Int. Conf. Comput. Vis. pp. 1854--1862 (2019)

\bibitem{liang2022balancing}
Liang, M., Zhou, J., Wei, W., Wu, Y.: Balancing between forgetting and acquisition in incremental subpopulation learning. In: European Conference on Computer Vision. pp. 364--380. Springer (2022)

\bibitem{liang2020instance}
Liang, S., Huang, Z., Liang, M., Yang, H.: Instance enhancement batch normalization: An adaptive regulator of batch noise. In: Proceedings of the AAAI Conference on Artificial Intelligence. vol.~34, pp. 4819--4827 (2020)

\bibitem{lin2014microsoft}
Lin, T.Y., Maire, M., Belongie, S., Hays, J., Perona, P., Ramanan, D., Doll{\'a}r, P., Zitnick, C.L.: Microsoft coco: Common objects in context. In: Eur. Conf. Comput. Vis. pp. 740--755 (2014)

\bibitem{luo2022few}
Luo, J., Si, W., Deng, Z.: Few-shot learning for radar signal recognition based on tensor imprint and re-parameterization multi-channel multi-branch model. IEEE Signal Processing Letters  \textbf{29},  1327--1331 (2022)

\bibitem{ma2018shufflenet}
Ma, N., Zhang, X., Zheng, H.T., Sun, J.: Shufflenet v2: Practical guidelines for efficient cnn architecture design. In: Proceedings of the European conference on computer vision (ECCV). pp. 116--131 (2018)

\bibitem{qin2021fcanet}
Qin, Z., Zhang, P., Wu, F., Li, X.: Fcanet: Frequency channel attention networks. In: Proceedings of the IEEE/CVF international conference on computer vision. pp. 783--792 (2021)

\bibitem{russakovsky2015imagenet}
Russakovsky, O., Deng, J., Su, H., Krause, J., Satheesh, S., Ma, S., Huang, Z., Karpathy, A., Khosla, A., Bernstein, M., et~al.: Imagenet large scale visual recognition challenge. International journal of computer vision  \textbf{115}(3),  211--252 (2015)

\bibitem{2017Grad}
Selvaraju, R.R., Cogswell, M., Das, A., Vedantam, R., Parikh, D., Batra, D.: Grad-cam: Visual explanations from deep networks via gradient-based localization. In: International Conference on Computer Vision (2017)

\bibitem{simonyan2014very}
Simonyan, K., Zisserman, A.: Very deep convolutional networks for large-scale image recognition. arXiv preprint arXiv:1409.1556  (2014)

\bibitem{touvron2022resmlp}
Touvron, H., Bojanowski, P., Caron, M., Cord, M., El-Nouby, A., Grave, E., Izacard, G., Joulin, A., Synnaeve, G., Verbeek, J., et~al.: Resmlp: Feedforward networks for image classification with data-efficient training. IEEE Transactions on Pattern Analysis and Machine Intelligence  (2022)

\bibitem{ulyanov2016instance}
Ulyanov, D., Vedaldi, A., Lempitsky, V.: Instance normalization: The missing ingredient for fast stylization. arXiv preprint arXiv:1607.08022  (2016)

\bibitem{vaswani2017attention}
Vaswani, A., Shazeer, N., Parmar, N., Uszkoreit, J., Jones, L., Gomez, A.N., Kaiser, {\L}., Polosukhin, I.: Attention is all you need. In: Adv. Neural Inform. Process. Syst. pp. 5998--6008 (2017)

\bibitem{wang2021recurrent}
Wang, J., Chen, Y., Yu, S.X., Cheung, B., LeCun, Y.: Recurrent parameter generators. arXiv preprint arXiv:2107.07110  (2021)

\bibitem{wang2020eca}
Wang, Q., Wu, B., Zhu, P., Li, P., Zuo, W., Hu, Q.: Eca-net: Efficient channel attention for deep convolutional neural networks. In: IEEE Conf. Comput. Vis. Pattern Recog. pp. 11534--11542 (2020)

\bibitem{wang2022repsr}
Wang, X., Dong, C., Shan, Y.: Repsr: Training efficient vgg-style super-resolution networks with structural re-parameterization and batch normalization. In: Proceedings of the 30th ACM International Conference on Multimedia. pp. 2556--2564 (2022)

\bibitem{woo2018cbam}
Woo, S., Park, J., Lee, J.Y., So~Kweon, I.: Cbam: Convolutional block attention module. In: Eur. Conf. Comput. Vis. pp. 3--19 (2018)

\bibitem{wu2018group}
Wu, Y., He, K.: Group normalization. In: Proceedings of the European conference on computer vision (ECCV). pp. 3--19 (2018)

\bibitem{yu2022metaformer}
Yu, W., Luo, M., Zhou, P., Si, C., Zhou, Y., Wang, X., Feng, J., Yan, S.: Metaformer is actually what you need for vision. In: Proceedings of the IEEE/CVF conference on computer vision and pattern recognition. pp. 10819--10829 (2022)

\bibitem{zhang2020resnest}
Zhang, H., Wu, C., Zhang, Z., Zhu, Y., Zhang, Z., Lin, H., Sun, Y., He, T., Mueller, J., Manmatha, R., et~al.: Resnest: Split-attention networks. arXiv preprint arXiv:2004.08955  (2020)

\bibitem{Zhang21}
Zhang, M., Yu, X., Rong, J., Ou, L.: Repnas: Searching for efficient re-parameterizing blocks. arXiv 2109.03508  (2021)

\bibitem{zhang2022cs}
Zhang, R., Wei, J., Lu, W., Zhang, L., Ji, Y., Xu, J., Lu, X.: Cs-rep: Making speaker verification networks embracing re-parameterization. In: ICASSP 2022-2022 IEEE International Conference on Acoustics, Speech and Signal Processing (ICASSP). pp. 7082--7086. IEEE (2022)

\bibitem{zhong2023esa}
Zhong, S., Huang, Z., Wen, W., Yang, Z., Qin, J.: Esa: Excitation-switchable attention for convolutional neural networks. Neurocomputing  \textbf{557},  126706 (2023)

\bibitem{zhong2022mix}
Zhong, S., Wen, W., Qin, J.: Mix-pooling strategy for attention mechanism. arXiv preprint arXiv:2208.10322  (2022)

\bibitem{zhong2023spem}
Zhong, S., Wen, W., Qin, J.: Spem: Self-adaptive pooling enhanced attention module for image recognition. In: International Conference on Multimedia Modeling. pp. 41--53. Springer (2023)

\bibitem{zhong2023lsas}
Zhong, S., Wen, W., Qin, J., Chen, Q., Huang, Z.: Lsas: Lightweight sub-attention strategy for alleviating attention bias problem. In: 2023 IEEE International Conference on Multimedia and Expo (ICME). pp. 2051--2056. IEEE (2023)

\bibitem{zhou2022cmb}
Zhou, H., Liu, L., Zhang, H., He, H., Zheng, N.: Cmb: A novel structural re-parameterization block without extra training parameters. In: 2022 International Joint Conference on Neural Networks (IJCNN). pp.~1--9. IEEE (2022)

\bibitem{zhu2019empirical}
Zhu, X., Cheng, D., Zhang, Z., Lin, S., Dai, J.: An empirical study of spatial attention mechanisms in deep networks. In: Int. Conf. Comput. Vis. pp. 6688--6697 (2019)

\end{thebibliography}
\end{document}